\pdfoutput=1 
\RequirePackage{silence}
\documentclass[a4paper,svgnames]{amsart}

\usepackage[
    hmarginratio={1:1},
    vmarginratio={1:1},
    lmargin=50.0pt,
    tmargin=45.0pt
]{geometry}

\usepackage{booktabs} 
\allowdisplaybreaks 

\usepackage{siunitx}
\usepackage[T1]{fontenc}
\usepackage{latexsym,exscale,amsfonts,amssymb,mathtools}
\usepackage{amsmath,amsthm,amsfonts,amssymb,amscd,textcomp,bbm}
\usepackage{mathrsfs,stackrel}
\DeclareMathAlphabet{\mathscrbf}{OMS}{mdugm}{b}{n}

\usepackage{etoolbox}
\usepackage{xparse}
\usepackage{aliascnt}
\usepackage{cite}

\usepackage{tikz,tikz-cd}
\usetikzlibrary{
    matrix,arrows.meta,decorations.markings,
    shapes.multipart,decorations.pathreplacing,
    decorations.pathmorphing,shapes.geometric
}
\usepackage{tikz-3dplot} 
\usepackage{dynkin-diagrams} 
\usepackage{ytableau} 
\usepackage{nicematrix} 
\usepackage{subcaption} 

\usepackage{array}
\newcolumntype{C}{>{$}c<{$}} 
\newcolumntype{P}[1]{>{\centering\arraybackslash}p{#1}} 

\usepackage{xcolor,colortbl}

\definecolor{lightgreen}{HTML}{CCFFCC} 
\definecolor{lightblue}{HTML}{CCCCFF} 

\definecolor{mygray}{gray}{0.6}
\definecolor{mygraydark}{gray}{0.4}
\definecolor{mygraylight}{gray}{0.85}

\definecolor{spinach}{RGB}{46,139,87}
\definecolor{tomato}{RGB}{255,99,71}
\definecolor{orchid}{RGB}{143,40,194}
\definecolor{neon}{RGB}{77,77,255} 
\definecolor{lightneon}{RGB}{110,110,255} 
\definecolor{pumpkin}{RGB}{224,180,80} 
\definecolor{citron}{RGB}{190,180,90} 
\definecolor{lava}{RGB}{207,16,32}
\definecolor{cream}{RGB}{255,253,208}
\definecolor{verdigris}{RGB}{67,179,174} 

\definecolor{mydarkblue}{RGB}{10,10,170}
\definecolor{darkspinach}{RGB}{20,70,20}
\definecolor{darktomato}{RGB}{155,40,30}
\definecolor{darkorchid}{RGB}{50,10,100}
\definecolor{darklava}{RGB}{150,8,16}

\definecolor{zero}{RGB}{0,0,0}
\definecolor{one}{RGB}{255,0,0}
\definecolor{two}{RGB}{0,255,0}
\definecolor{three}{RGB}{0,0,255}

\usepackage{todonotes}

\usepackage{enumitem}
\setlist[enumerate]{itemsep=0.15cm,label=\emph{\upshape(\arabic*)}}
\setlist[enumerate,2]{itemsep=0.15cm,label=\emph{\upshape(\roman*)}}
\setlist[enumerate,3]{itemsep=0.15cm,label=\emph{\upshape(\Alph*)}}

\let\emph\relax
\DeclareTextFontCommand{\emph}{\bfseries\em}

\renewcommand{\dots}{\text{...}}

\newcommand{\mystrut}{\rule[-0.2\baselineskip]{0pt}{0.9\baselineskip}}

\usepackage[all]{xy}
\usepackage{tikz}
\usetikzlibrary{
  cd,
  decorations,
  decorations.markings,
  decorations.pathreplacing,
  decorations.pathmorphing,
  arrows.meta,
  shapes,
  shapes.callouts,
  positioning,
  matrix,
  calc
}

\tikzset{
  anchorbase/.style={baseline={([yshift=-0.5ex]current bounding box.center)}},
  tinynodes/.style={font=\tiny,text height=0.25ex,text depth=0.05ex},
  smallnodes/.style={font=\scriptsize,text height=0.75ex,text depth=0.15ex},
  usual/.style={line width=2.0,color=black},
  crossline/.style={preaction={draw=white,line width=5.75pt,-}}
}

\tikzset{
  directed/.style={postaction={decorate,decoration={markings,
    mark=at position #1 with {\arrow[line width=0.3mm, black]{>}}}}}
}

\tikzset{
  startstop/.style={ellipse, draw, fill=blue!20, text width=2cm, text centered, minimum height=1cm},
  process/.style={rectangle, draw, fill=orange!20, text width=2.5cm, text centered, minimum height=1cm},
  arrow/.style={thick, -{Stealth[length=3mm]}}
}

\newcommand{\C}{\mathbb{C}}

\def\NewTheorem#1{%
    \newaliascnt{#1}{equation}%
    \newtheorem{#1}[#1]{#1}%
    \aliascntresetthe{#1}%
    \expandafter\def\csname #1autorefname\endcsname{#1}%
}

\def\equationautorefname~#1\null{(#1)\null}

\numberwithin{equation}{subsection}

\NewTheorem{Proposition}
\NewTheorem{Theorem}
\NewTheorem{Corollary}
    \AtEndEnvironment{Corollary}{\null\hfill$\square$}%
\NewTheorem{Lemma}
\NewTheorem{Conjecture}
\NewTheorem{Speculation}
\NewTheorem{Observation} 
\NewTheorem{Assumption} 

\theoremstyle{definition}
\NewTheorem{Definition}
    \AtEndEnvironment{Definition}{\null\hfill$\Diamond$}%
\NewTheorem{Classification Problem} 
    \AtEndEnvironment{Classification Problem}{\null\hfill$\Diamond$}%
\NewTheorem{Notation}
    \AtEndEnvironment{Notation}{\null\hfill$\Diamond$}%
\NewTheorem{Example}
    \AtEndEnvironment{Example}{\null\hfill$\Diamond$}%
\NewTheorem{Examples}
    \AtEndEnvironment{Examples}{\vskip-10mm\null\hfill$\Diamond$}%

\theoremstyle{remark}
\NewTheorem{Remark}
    \AtEndEnvironment{Remark}{\null\hfill$\Diamond$}%
\NewTheorem{Question}
    \AtEndEnvironment{Question}{\null\hfill$\Diamond$}%

\usepackage[hypertexnames=false]{hyperref}
\usepackage{bookmark}
\hypersetup{
    pdftoolbar=true,
    pdfmenubar=true,
    pdffitwindow=false,
    pdfstartview={FitH},
    pdftitle={On knot detection via picture recognition},
    pdfauthor={Anne Dranowski, Yura Kabkov and Daniel Tubbenhauer},
    pdfsubject={},
    pdfcreator={Anne Dranowski, Yura Kabkov and Daniel Tubbenhauer},
    pdfproducer={Anne Dranowski, Yura Kabkov and Daniel Tubbenhauer},
    pdfkeywords={},
    pdfnewwindow=true,
    colorlinks=true,
    linkcolor=mydarkblue,
    citecolor=teal,
    filecolor=magenta,
    urlcolor=orchid,
    linkbordercolor=lava,
    citebordercolor=teal,
    urlbordercolor=orchid,
    linktocpage=true
}
\usepackage[nameinlink, noabbrev,capitalize]{cleveref}

\usepackage{float} 
\usepackage[justification=centering]{caption}
\def\makeautorefname#1#2{\csdef{#1autorefname}{#2}}
\makeautorefname{section}{Section} %
\makeautorefname{subsection}{Section} %
\makeautorefname{subsubsection}{Section} %

\begin{document}

\title[On knot detection via picture recognition]{On knot detection via picture recognition}
\author[A. Dranowski, Y. Kabkov and D. Tubbenhauer]{Anne Dranowski, Yura Kabkov and Daniel Tubbenhauer}

\address{A.D.: annedranowski@gmail.com} 

\address{Y.K.: yura.yulias.dream.prog@gmail.com} 

\address{D.T.: The University of Sydney, School of Mathematics and Statistics F07, Office Carslaw 827, NSW 2006, Australia, \href{http://www.dtubbenhauer.com}{www.dtubbenhauer.com}, \href{https://orcid.org/0000-0001-7265-5047}{ORCID 0000-0001-7265-5047}}
\email{daniel.tubbenhauer@sydney.edu.au}

\begin{abstract}
Our goal is to one day take a photo of a knot and have a phone automatically recognize it. In this expository work, we explain a strategy to approximate this goal, using a mixture of modern machine learning methods 
(in particular convolutional neural networks and transformers for image recognition)
and traditional algorithms (to compute quantum invariants like the Jones polynomial).
We present simple baselines that predict crossing number directly from images, showing that even lightweight CNN and transformer architectures can recover meaningful structural information. 
The longer-term aim is to combine these perception modules with symbolic reconstruction into planar diagram (PD) codes, enabling downstream invariant computation for robust knot classification. 
This two-stage approach highlights the complementarity between machine learning, which handles noisy visual data, and invariants, which enforce rigorous topological distinctions.
\end{abstract}


\addtocontents{toc}{\protect\setcounter{tocdepth}{1}}

\maketitle
\tableofcontents

\section{Introduction}\label{S:Intro}

We explain why and how a combination of modern machine learning techniques and traditional algorithms offers a promising route to automated knot recognition. Our guiding principle is simple:
\begin{gather*}
\fcolorbox{orchid!50}{spinach!10}{\mystrut``Recognize first from the image, then compute invariants.''} 
\end{gather*}
That is, begin with image-based prediction, then refine and verify using topological invariants.
This image-first strategy, followed by the calculation of 
invariants, appears to be new and forms the foundation of our approach.

\subsection{The paper in a nutshell} 

\begin{figure}[ht]
\centering
\begin{subfigure}[b]{0.45\textwidth}
\centering
\includegraphics[height=4.5cm]{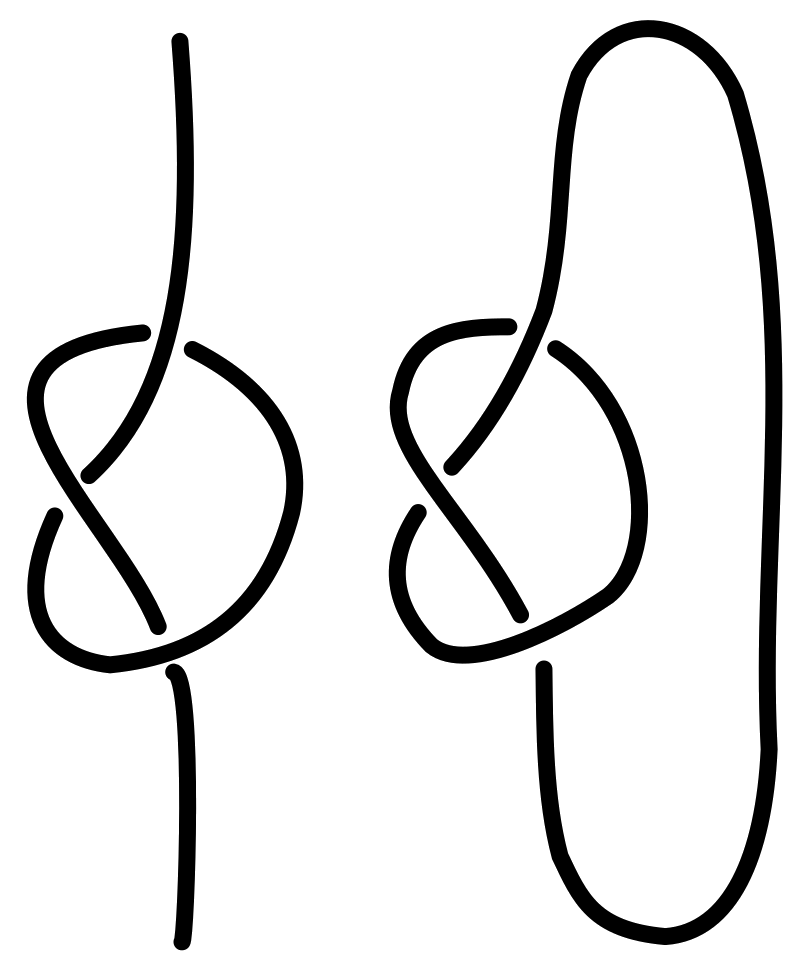}
\caption{A trefoil knot.}
\label{fig:trefoil}
\end{subfigure}
\hfill
\begin{subfigure}[b]{0.45\textwidth}
\centering
\includegraphics[height=4.5cm]{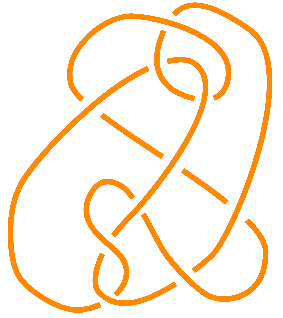}
\caption{An unknot in disguise.}
\label{fig:other}
\end{subfigure}
\caption{In knot theory, a knot is a closed loop rather than a strand with free ends, since loose ends could simply be undone. 
(Some of the knots we discuss, such as protein knots, are not knots in this mathematical sense, but we will ignore the difference.)
In \autoref{fig:other} we have a diagram of the unknot that looks anything but trivial.  
{\tiny Pictures from \url{https://en.wikipedia.org/wiki/Trefoil\_knot} and \cite{MR3193721}.}}
\label{Fig:Trefoil}
\end{figure}

Knot recognition, the process of describing the knot we are looking at, is surprisingly hard.  
Knots, like cats, can be classified according to distinguishing attributes: just as a Siamese is distinct from a Persian, so is the unknot (an unknotted string) distinct from the trefoil, cf.~\autoref{Fig:Trefoil}.  
The difficulty is that, unlike cats, we have little intuitive training data for knots, and diagrams of the same knot can vary wildly in appearance.

\begin{figure}[tbp]
\centering
\begin{subfigure}[t]{0.44\textwidth}
\centering
\includegraphics[height=4.5cm]{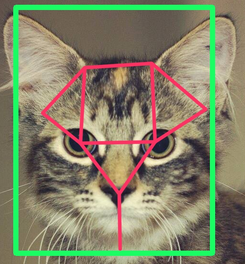}
\caption{A cat.}
\label{fig:pair1}
\end{subfigure}
\hfill
\begin{subfigure}[t]{0.52\textwidth}
\centering
\includegraphics[height=4.5cm]{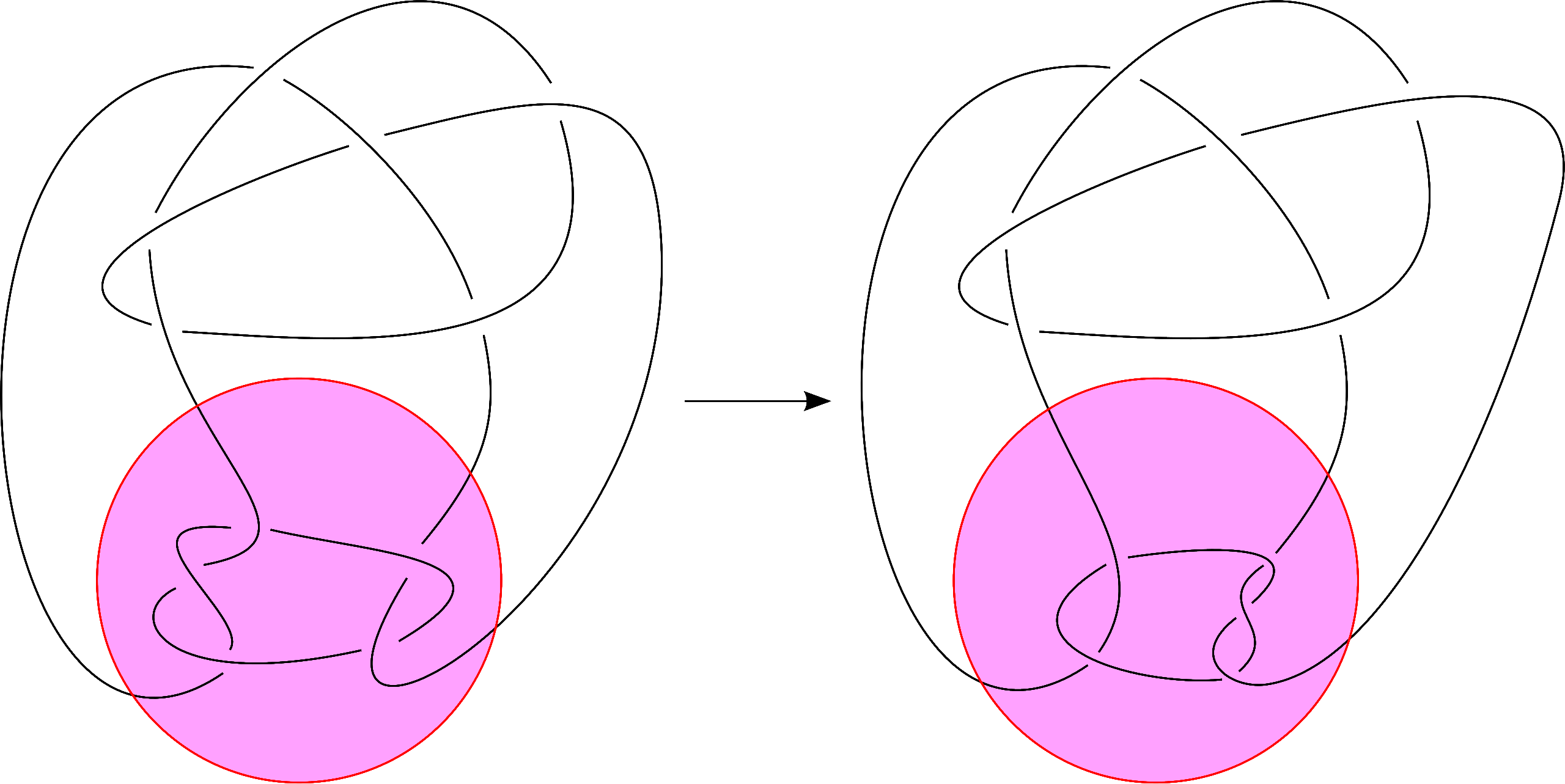}
\caption{The Kinoshita--Terasaka (left) and Conway (right) knots.}
\label{fig:pair2}
\end{subfigure}
\caption{\autoref{fig:pair1} shows how a cat can be detected locally, while \autoref{fig:pair2} shows two knots that look almost identical.  
{\tiny Pictures from somewhere (and then marked) and \url{https://en.wikipedia.org/wiki/Kinoshita-Terasaka_knot}.}}
\label{fig:Pair}
\end{figure}

\begin{figure}[ht]
\centering
\includegraphics[height=4.5cm]{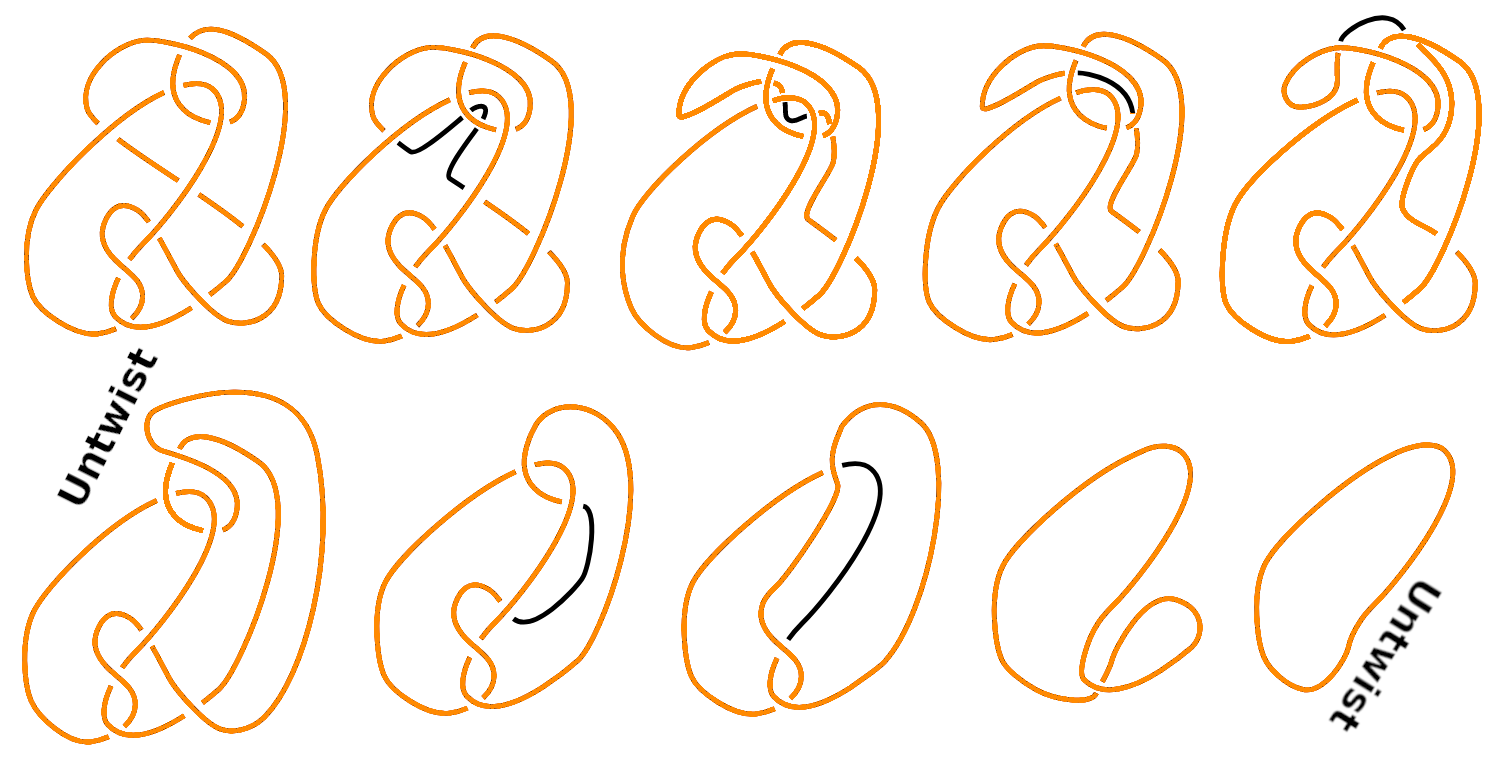}
\caption{The right knot in \autoref{Fig:Trefoil} is actually not knotted at all.  
{\tiny Picture from \cite{MR3193721}.}}
\label{Fig:Culprit}
\end{figure}

Three obstacles stand out (see also \autoref{SS:PictureRecognition}):
\begin{enumerate}
\item \emph{We do not commonly encounter knots}, nor learn to recognize them visually.
Humans can memorize cat breeds by many cues (fur, face, ears), but parsing a knot requires tracking crossings and over/under structure.  
The Kinoshita--Terasaka and Conway knots in \autoref{fig:pair2} are a classic example of two distinct knots that look essentially identical.

\item \emph{Knots often appear in disguise}: Your computer cables may look knotted but can be undone.
Many complicated diagrams are actually the unknot, see e.g. \autoref{Fig:Culprit}, and similarly all knots have ``bad'' diagrams. Detecting a knot from a bad diagram is especially challenging.

\item \emph{No local giveaway}: Cats can be recognized from e.g. the local patterns of their face, cf.\ \autoref{fig:pair1}; knots have no such signature: two diagrams of the same knot can be locally unrelated or two diagrams that are mostly the same can be different knots, cf.\ \autoref{Fig:Culprit} and \autoref{fig:pair2}. This lack of a local signature means global reasoning is essential.
\end{enumerate}

\begin{figure}[ht]
\centering
\begin{subfigure}[b]{0.45\textwidth}
\centering
\includegraphics[height=4.5cm]{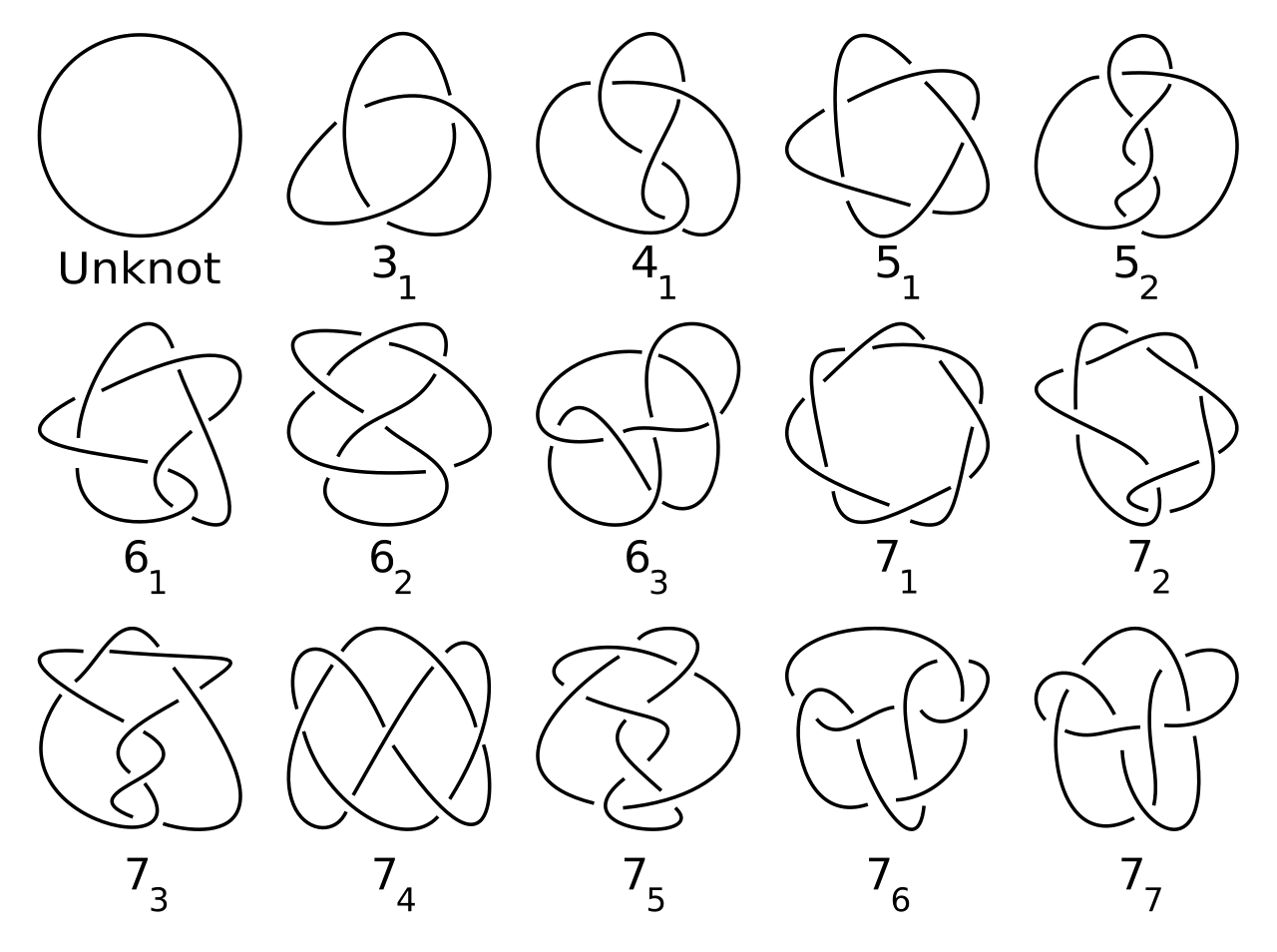}
\caption{A list of mathematical knots.}
\label{fig:DNA1}
\end{subfigure}
\hfill
\begin{subfigure}[b]{0.45\textwidth}
\centering
\includegraphics[height=4.5cm]{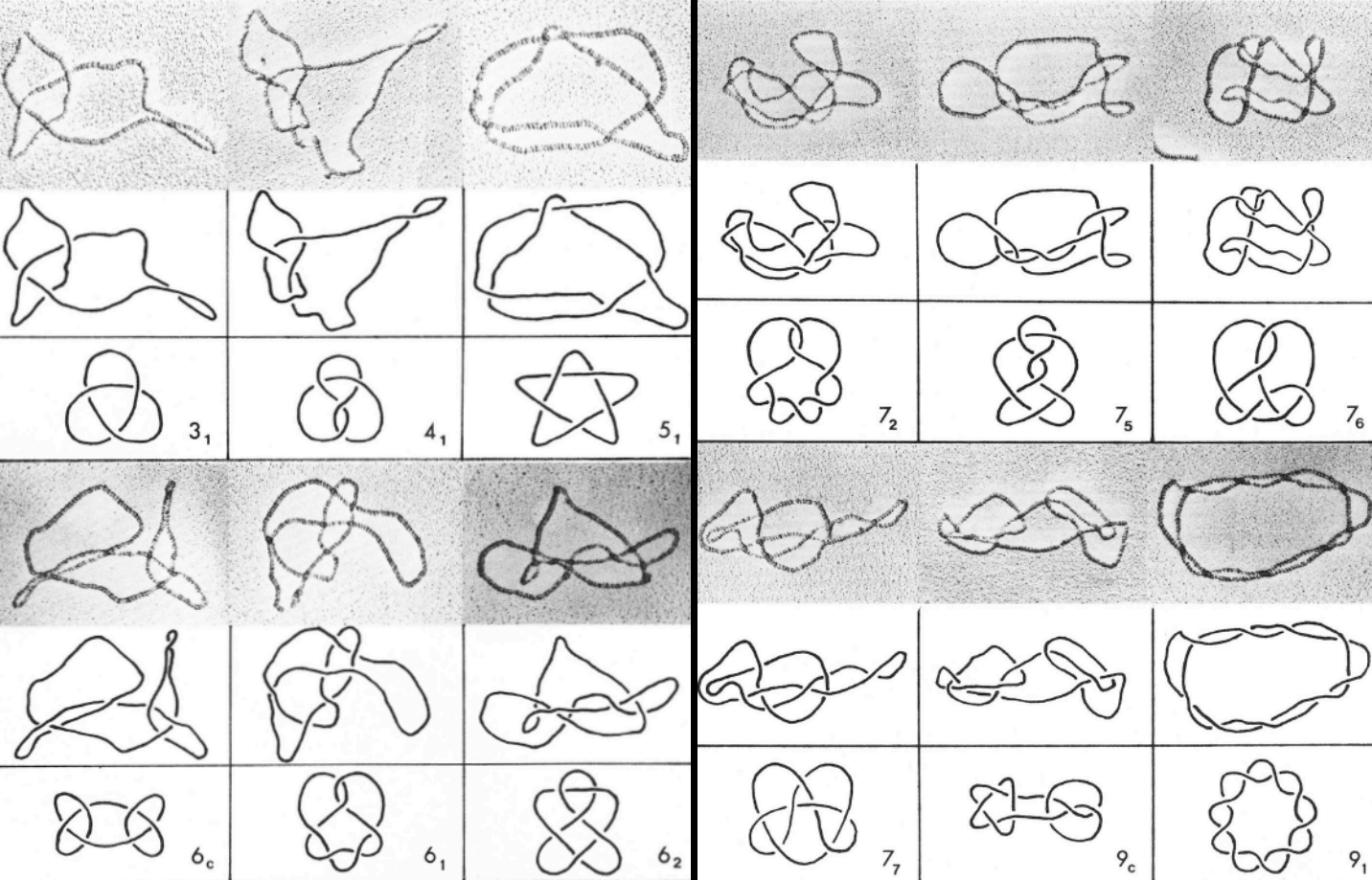}
\caption{A list of DNA knots.}
\label{fig:DNA2}
\end{subfigure}
\caption{Top row in \autoref{fig:DNA2}: microscopy images of DNA knots; middle: human sketches; bottom: matches to known knots as in \autoref{fig:DNA1}.  
Wouldn’t it be better to have this automated?  
{\tiny Pictures from \url{https://en.wikipedia.org/wiki/Prime\_knot} and \cite{Dean1985}.}}
\label{Fig:DNA}
\end{figure}

Why bother?  
Knots are not just mathematical curiosities, they appear 
in DNA topology, protein folding, polymer science, fluid dynamics, and beyond.  
In these fields, identifying the knot from an image (for instance, from an electron microscope) is a common task, cf.~\autoref{Fig:DNA}.  
At present this is often done by hand: drawing a diagram from the image, then matching it against known knot types.  
This process is labor-intensive, error-prone, and strongly dependent on expert intuition. Automating it would improve speed, reproducibility, and accessibility.

Our long-term goal is to take a picture of a knot and have a computer tell us what it is: think about an app on your phone.  
In this paper we take a tiny first step: predicting the number of crossings of a knot diagram directly from an image using convolutional neural networks (CNNs) and transformers.  
Ultimately, we want the model to output a planar diagram (PD) presentation so that established algorithms can compute quantum invariants such as the Jones polynomial.  
This two-stage approach balances the strengths of modern machine learning with the reliability of mathematical invariants.

Two observations shape our strategy:
\begin{enumerate}[resume]
\item Most knots seen in real images are \emph{small}, in the sense of having low minimal crossing number; due to physical constraints, tying habits, and biases in natural examples, but this is also a general phenomena in random samples of knots.
\item Many quantum invariants, while theoretically limited, work \emph{extremely well} on \emph{small} knots, making them powerful once a reasonable PD presentation is recovered.
\end{enumerate}
With these observations in mind, we build a pipeline that learns from images, reconstructs a diagram, and then applies invariants for final classification.  
The rest of the paper develops this approach and presents some first results toward automatic knot recognition. 

As we will explain:
\begin{gather*}
\fcolorbox{orchid!50}{spinach!10}{\mystrut``We expect this to work well for knot diagrams up to $30$ crossings.''} 
\end{gather*}

\subsection{Knot theory and machine learning (ML)}

Machine learning has so far been applied to knot theory primarily in settings where the input is a \emph{symbolic or geometric encoding} of the knot, such as polynomials, braid words, or spatial graphs, rather than \emph{raw images}. 
Most references originating within knot theory itself \cite{MR4101599,MR4316400,Davies2021,Gukov2021,MR4555584,Gukov2024a} focus on pattern prediction of various invariants, typically using braid words or other algebraic encodings as input. 
In contrast, biology-inspired approaches tend to work with geometric inputs and achieve highly accurate predictions of knottedness, though usually restricted to ``very small'' knots (around six crossings) \cite{Jumper2021,Silva2024}. 
To the best of our knowledge, data-driven approaches that extend to knots with up to about $20$ crossings are rare, with only a few recent efforts in this direction \cite{DlGuSa-data,TuZh-knot-data,LaTuVaZh-big-data-2}. 
It is natural to argue that the real difficulty of knot recognition begins only for larger knots, well beyond the range that most existing studies address.

\emph{What is not common: image-first recognition.}  
To the best of our knowledge, there is virtually no prior work that starts from photographs or arbitrary two-dimensional raster images of knots, aims to output a PD presentation, or produces a knot label based on topological equivalence under Reidemeister moves. While some computer-vision applications tackle ropes and knots for non-topological goals (e.g. damage detection or task-driven knot-tying), none produce PD presentations usable for invariant computations. This gap motivates our
``image\,$\longmapsto$ PD $\longmapsto$ invariant'' pipeline:
\begin{gather*}
\fcolorbox{orchid!50}{spinach!10}{\mystrut$\begin{tikzpicture}[anchorbase]
\node at (0,0) {\reflectbox{\includegraphics[height=2cm]{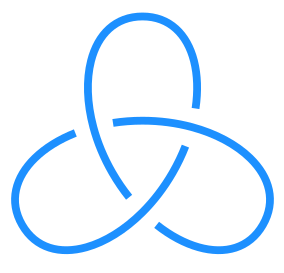}}};
\end{tikzpicture}
\longmapsto
PD[X[1, 5, 2, 4],X[3, 1, 4, 6],X[5, 3, 6, 2]]
\longmapsto
-q^{-4}+q^{-3}+q^{-1}\longmapsto\text{trefoil}$}
\end{gather*}
Where prior work applies ML either to symbolic invariant prediction or diagram-based encoding, our work proposes a new direction: an image-first pathway that integrates perception (segmenting and interpreting the knot), topology (PD presentation reconstruction), and algebra (invariants) into a unified end-to-end pipeline.

\subsection{Summary}

In this paper we:
\begin{enumerate}[resume]
\item formalize \emph{visual knot recognition} as an ``image\,$\longmapsto$ PD $\longmapsto$ invariant'' pipeline and fix PD presentations as the interface between vision and topology;
\item explain \emph{why this should work in practice}: most knots that appear in images are small (low minimal crossing number), and classical invariants are exceptionally effective on small knots;
\item set out design principles (topological invariance, global consistency, interpretability) that guide our models and evaluation;
\item implement first baselines that predict the number of crossings in images using, and comparing, a vanilla neural network (NN), a convolutional NN (CNN), and a convolutional vision transformer (CvT), as a step toward full PD presentation recovery.
\end{enumerate}
Our workflow is illustrated below.
\begin{gather}\label{Eq:Flow}
\text{This work: }\;
\scalebox{0.7}{\begin{tikzpicture}[anchorbase,node distance=0.5cm]
\node (start) [startstop] {Input Image};
\node (pre)   [process, below=of start] {Pre-processing\\(resize, denoise)};
\node (conv)  [process, below=of pre, inner sep=0] {NN Layers\\(e.g.\ CNN)};
\node (out)   [startstop, below=of conv, inner sep=0] {Output:\\ Crossing count};
\draw [arrow] (start) -- (pre);
\draw [arrow] (pre)   -- (conv);
\draw [arrow] (conv)  -- (out);
\end{tikzpicture}}
\quad,\quad
\text{Future work: }\;
\scalebox{0.7}{\begin{tikzpicture}[anchorbase,node distance=0.5cm]
\node (start) [startstop] {Input Image};
\node (pre)   [process, below=of start] {Pre-processing\\(resize, denoise)};
\node (conv)  [process, below=of pre, inner sep=0] {NN Layers\\(e.g.\ CNN)};
\node (pd)    [startstop, below=of conv, inner sep=0] {Output:\\ PD presentation};
\node (inv)   [process, below=of pd, text width=4cm, minimum width=4cm, inner sep=0] {Compute invariant\\(e.g.\ Jones polynomial)};
\node (lab)   [startstop, below=of inv, inner sep=0] {Output:\\ Knot label};
\draw [arrow] (start) -- (pre);
\draw [arrow] (pre)   -- (conv);
\draw [arrow] (conv)  -- (pd);
\draw [arrow] (pd)    -- (inv);
\draw [arrow] (inv)   -- (lab);
\end{tikzpicture}}
\end{gather}
For readers concerned about the ``undoing'' problem highlighted in \autoref{Fig:Culprit}: \autoref{S:Strategy} explains why the right-hand pipeline above is feasible, at least probabilistically, given the small-knot bias in real images and the behavior of invariants in that regime.

\emph{Why not ``just compute the Jones polynomial from the pixels''?} There is strong complexity-theoretic evidence that approximating the Jones polynomial is very hard (see e.g. \cite{MR3354608} for precise statements), so a reliable classical direct-from-image surrogate is unlikely to exist. Our pipeline isolates perception, where errors are measurable and correctable, and then hands a valid combinatorial object to the invariant engine; in that regime computing Jones is fast for the small-crossing cases we actually see, and standard invariants distinguish those knots unusually well. In short: perceive first, then compute invariants yields stability, interpretability, and practical runtime where a direct numerical ``Jones-from-pixels'' route would be brittle.

\subsection{Remarks}

We park the following here:

\begin{Remark}
Our paper is mostly self-contained, but the reader is expected to have some basic background in knot theory and deep learning techniques. Standard references for these include \cite{Ad-knots} or \cite{GoBeCo-deep-learning}. Not much is needed about quantum invariants (the part on the Jones polynomial in \cite{Ad-knots} should be sufficient), but our conventions are taken from \cite{Tu-qt}.
\end{Remark}

\begin{Remark}
The reader needs to be a bit careful with the difference between the crossing number, which is the minimal number of crossings needed to draw a given knot, and the number of crossings of a diagram. We predict the latter.
\end{Remark}

\begin{Remark}
The paper is recommended to be read in color. 
\end{Remark}

\begin{Remark}
All data files associated to this paper can be found online: 
\url{https://github.com/annedranowski/knot-cnn}. 
\end{Remark}

\noindent\textbf{Acknowledgments.}
We would like to thank Radmila Sazdanovic for directing us to the work of Robert Scharein (see \cite{knotplot} for striking visualizations of knots) on a knot detection app, which, unfortunately, we were unable to locate. We are especially grateful to Saul Schleimer, who went above and beyond in their role as an arXiv moderator by carefully reading a submitted version of our manuscript and pointing out an error. (Thanks to Saul's close attention, we were able to correct the issue before the paper was officially posted.) This work was carried out through the PRIMES program at MIT, under the Yulia's Dream track, with the generous support of the organizers. DT reflects that the pyramids will likely outlast humanity, while knot detection collapses quickly beyond the easiest cases.

\section{Knot detection via image recognition}\label{S:KnotDetection}

In this section we explain the problem, knot recognition, and our solution strategy.

\subsection{The problem mathematically: detecting knots}\label{SS:MathDetection}

We start by recalling two central decision problems in knot theory: recognizing the unknot and determining equivalence between two knots. The latter is our main focus, but the former has more available results.

To get started, recall that a \emph{knot diagram} is a projection of an embedding \(K: S^1\hookrightarrow \mathbb{R}^3\), presented as a 4-regular planar graph with crossing information. For example:
\begin{gather*}
\begin{tikzpicture}[anchorbase,scale=1]
	\draw[usual] (0.5,0) to[out=90,in=270] (0,0.5);
	\draw[usual] (0,0) to[out=90,in=270] (0.5,0.5);
	\draw[usual] (1,0) to[out=90,in=270] (1,0.5);
	\draw[usual,fill=cream] (0.25,0.25) circle (0.2cm);
	\draw[usual] (0,0.5) to[out=90,in=270] (0,1);
	\draw[usual] (0.5,0.5) to[out=90,in=270] (1,1);
	\draw[usual] (1,0.5) to[out=90,in=270] (0.5,1);
	\draw[usual,fill=cream] (0.75,0.75) circle (0.2cm);
	\draw[usual] (0.5,1) to[out=90,in=270] (0,1.5);
	\draw[usual] (0,1) to[out=90,in=270] (0.5,1.5);
	\draw[usual] (1,1) to[out=90,in=270] (1,1.5);
	\draw[usual,fill=cream] (0.25,1.25) circle (0.2cm);
	\draw[usual] (0,1.5) to[out=90,in=270] (0,2);
	\draw[usual] (0.5,1.5) to[out=90,in=270] (1,2);
	\draw[usual] (1,1.5) to[out=90,in=270] (0.5,2);
	\draw[usual,fill=cream] (0.75,1.75) circle (0.2cm);
	\draw[usual] (1,2) to[out=90,in=180] (1.25,2.25) to[out=0,in=90] (1.5,2) to (1.5,0) to[out=270,in=0] (1.25,-0.25) to[out=180,in=270] (1,0);
	\draw[usual] (0.5,2) to[out=90,in=180] (1.25,2.5) to[out=0,in=90] (2,2) to (2,0) to[out=270,in=0] (1.25,-0.5) to[out=180,in=270] (0.5,0);
	\draw[usual] (0,2) to[out=90,in=180] (1.25,2.75) to[out=0,in=90] (2.5,2) to (2.5,0) to[out=270,in=0] (1.25,-0.75) to[out=180,in=270] (0,0);
\end{tikzpicture}
\xrightarrow[\text{vertex size}]{\text{decrease}}
\begin{tikzpicture}[anchorbase,scale=1]
\draw[usual] (0.5,0) to[out=90,in=270] (0,0.5);
\draw[usual] (0,0) to[out=90,in=270] (0.5,0.5);
\draw[usual] (1,0) to[out=90,in=270] (1,0.5);
\draw[usual] (0,0.5) to[out=90,in=270] (0,1);
\draw[usual] (0.5,0.5) to[out=90,in=270] (1,1);
\draw[usual] (1,0.5) to[out=90,in=270] (0.5,1);
\draw[usual] (0.5,1) to[out=90,in=270] (0,1.5);
\draw[usual] (0,1) to[out=90,in=270] (0.5,1.5);
\draw[usual] (1,1) to[out=90,in=270] (1,1.5);
\draw[usual] (0,1.5) to[out=90,in=270] (0,2);
\draw[usual] (0.5,1.5) to[out=90,in=270] (1,2);
\draw[usual] (1,1.5) to[out=90,in=270] (0.5,2);
\draw[usual] (1,2) to[out=90,in=180] (1.25,2.25) to[out=0,in=90] (1.5,2) to (1.5,0) to[out=270,in=0] (1.25,-0.25) to[out=180,in=270] (1,0);
\draw[usual] (0.5,2) to[out=90,in=180] (1.25,2.5) to[out=0,in=90] (2,2) to (2,0) to[out=270,in=0] (1.25,-0.5) to[out=180,in=270] (0.5,0);
\draw[usual] (0,2) to[out=90,in=180] (1.25,2.75) to[out=0,in=90] (2.5,2) to (2.5,0) to[out=270,in=0] (1.25,-0.75) to[out=180,in=270] (0,0);
\end{tikzpicture}
\xrightarrow[\text{crossings}]{\text{add}}
\begin{tikzpicture}[anchorbase,scale=1]
\draw[usual] (0.5,0) to[out=90,in=270] (0,0.5);
\draw[usual,crossline] (0,0) to[out=90,in=270] (0.5,0.5);
\draw[usual] (1,0) to[out=90,in=270] (1,0.5);
\draw[usual] (0,0.5) to[out=90,in=270] (0,1);
\draw[usual] (0.5,0.5) to[out=90,in=270] (1,1);
\draw[usual,crossline] (1,0.5) to[out=90,in=270] (0.5,1);
\draw[usual] (0.5,1) to[out=90,in=270] (0,1.5);
\draw[usual,crossline] (0,1) to[out=90,in=270] (0.5,1.5);
\draw[usual] (1,1) to[out=90,in=270] (1,1.5);
\draw[usual] (0,1.5) to[out=90,in=270] (0,2);
\draw[usual] (0.5,1.5) to[out=90,in=270] (1,2);
\draw[usual,crossline] (1,1.5) to[out=90,in=270] (0.5,2);
\draw[usual] (1,2) to[out=90,in=180] (1.25,2.25) to[out=0,in=90] (1.5,2) to (1.5,0) to[out=270,in=0] (1.25,-0.25) to[out=180,in=270] (1,0);
\draw[usual] (0.5,2) to[out=90,in=180] (1.25,2.5) to[out=0,in=90] (2,2) to (2,0) to[out=270,in=0] (1.25,-0.5) to[out=180,in=270] (0.5,0);
\draw[usual] (0,2) to[out=90,in=180] (1.25,2.75) to[out=0,in=90] (2.5,2) to (2.5,0) to[out=270,in=0] (1.25,-0.75) to[out=180,in=270] (0,0);
\end{tikzpicture}
\,.
\end{gather*}
This information is taken modulo planar isotopy and the Reidemeister moves (1,2 and 3):
\begin{gather*}
\text{$1$:}
\begin{tikzpicture}[anchorbase,scale=1]
\draw[usual,crossline] (0.5,0.75) to[out=270,in=0] (0.25,0.5) to[out=180,in=270] (0,1.25) to (0,1.5);
\draw[usual,crossline] (0,0) to (0,0.25) to[out=90,in=180] (0.25,1) to[out=0,in=90] (0.5,0.75);
\end{tikzpicture}
=
\begin{tikzpicture}[anchorbase,scale=1]
\draw[usual] (0,0) to[out=90,in=270] (0,1.5);
\end{tikzpicture}
=
\begin{tikzpicture}[anchorbase,scale=1]
\draw[usual,crossline] (0,0) to (0,0.25) to[out=90,in=180] (0.25,1) to[out=0,in=90] (0.5,0.75);
\draw[usual,crossline] (0.5,0.75) to[out=270,in=0] (0.25,0.5) to[out=180,in=270] (0,1.25) to (0,1.5);
\end{tikzpicture}
,\quad
\text{$2$:}
\begin{tikzpicture}[anchorbase,scale=1]
\draw[usual] (0.5,0) to[out=90,in=270] (0,0.75) to[out=90,in=270] (0.5,1.5);
\draw[usual,crossline] (0,0) to[out=90,in=270] (0.5,0.75) to[out=90,in=270] (0,1.5);
\end{tikzpicture}
=
\begin{tikzpicture}[anchorbase,scale=1]
\draw[usual] (0,0) to[out=90,in=270] (0,1.5);
\draw[usual] (0.5,0) to[out=90,in=270] (0.5,1.5);
\end{tikzpicture}
,\quad
\text{$3$:}
\begin{tikzpicture}[anchorbase,scale=1]
\draw[usual] (1,0) to[out=90,in=270] (0,1.5);
\draw[usual,crossline] (0.5,0) to[out=90,in=270] 
(0,0.75) to[out=90,in=270] (0.5,1.5);
\draw[usual,crossline] (0,0) to[out=90,in=270] (1,1.5);
\end{tikzpicture}
=
\begin{tikzpicture}[anchorbase,scale=1]
\draw[usual] (1,0) to[out=90,in=270] (0,1.5);
\draw[usual,crossline] (0.5,0) to[out=90,in=270] 
(1,0.75) to[out=90,in=270] (0.5,1.5);
\draw[usual,crossline] (0,0) to[out=90,in=270] (1,1.5);
\end{tikzpicture}
.
\end{gather*}

A \emph{knot} is then an equivalence class of knot diagrams, cf.\ \autoref{Fig:Culprit}, in particular, one knot is represented by many diagrams. A \emph{link} is a multiple component version, and knots and links can (often) be studied in tandem, and we will use links as such. 

The \emph{unknotting problem (UP)} asks: given a knot diagram \(D\), is \(D\) (a representative for) the unknot? More generally, the \emph{knottedness problem (KP)} considers two diagrams \(D_1, D_2\) and asks whether they represent the same knot.
Note that UP is a subproblem of KP, and since more is known about it, we quickly discuss it first.

A priori it is not clear whether there is any algorithm to decide UP. For example, if one takes the number of crossings of a diagram as a complexity measure (which we use throughout), then it
can happen that one needs to make a diagram more complex before one can simplify it, cf.\ \autoref{Fig:Culprit}. But it turns out that UP is decidable in a formal sense. 
%
%
%
%
In summary, UP is now known to lie in \(\mathrm{NP}\cap \mathrm{coNP}\) \cite{MR141106,MR1693203,MR4274879}, a tantalizing position often thought to be a prelude to polynomial-time algorithms; though whether it lies in \(\mathrm{P}\) is unknown while writing this paper. So one can think of UP as a \emph{semi-hard problem}.

In contrast, KP remains decidable but poorly understood in complexity terms. Haken's ``normal surface'' algorithms extend to decide equivalence, but imply a tower-of-exponentials time bound, as follows from \cite{MR1693203}. Moreover, reductions show that KP is at least NP-hard (and plausibly BQP-hard), while the best known upper classification is that it belongs to the elementary recursive class, a non-elementary but decidable class. This problem is computationally intractable: algorithmic bounds are colossal, and it is likely as hard as any problem in NP, or even quantum-hard. So one can think of KP as a \emph{really hard problem}.

When thinking about giving approximate answers, which is what NNs are good at,
UP and KP are in a decision landscape, which makes ``approximation'' subtle: approximations of a 0-vs-1 decision problem are typically vacuous. 
A standard remedy, familiar from approximation and parameterized complexity, is to study optimization surrogates whose optima certify (or strongly correlate with) detection/+decision, in our case, detection. 
For example, we use the crossing number as a surrogate complexity measure for the KP, cf.\ \autoref{S:CrossingDetection}.
As a very closely related analogy,
\cite{MR3017916} shows that, unless $\mathrm{P}=\mathrm{NP}$, there exists a constant $c_0>1$ such that no polynomial-time $c_0$-approximation exists for the computation of the graph crossing number (the minimal number of crossings one needs to draw a graph in the plane).

Hence, since KP is harder than UP and (probably) more difficult to approximate than the graph crossing number:
\begin{gather*}
\fcolorbox{orchid!50}{spinach!10}{\mystrut``Knot detection is a difficult problem, even for approximate solutions.''} 
\end{gather*}

\subsection{The problem as a test case for picture recognition: detecting knots}\label{SS:PictureRecognition}

We now formalize the task ``recognize knots from pictures''. Let $\mathcal{I}$ denote the space of images (grayscale or RGB or RGB--D) of a single, slender deformable linear object (e.g.\ a rope) placed in front of a static background under generic illumination with crossing information. For example:
\begin{gather}\label{Eq:Knot}
\begin{tikzpicture}[anchorbase]
\node at (0,0)
{\reflectbox{\includegraphics[height=4cm]{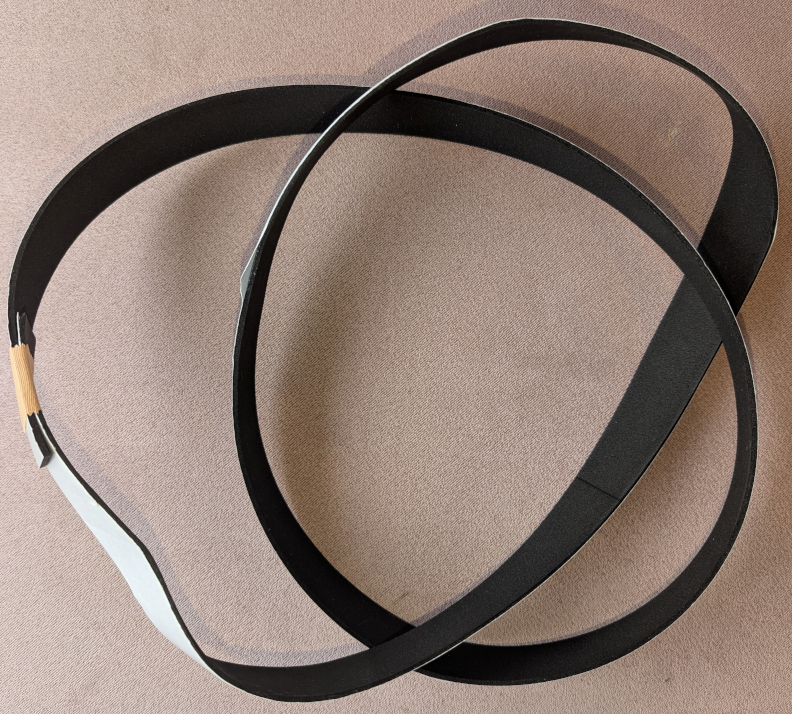}}};
\end{tikzpicture}
\end{gather}
Here, and without loss of (too much) generality, a single image generically produces a projection of the knot with finitely many transverse double points. In other words, we get a knot diagram, and we can think of $\mathcal{I}$ as the set of images of knot diagrams. 

Let $\mathcal{K}$ be the set of knots (equivalence classes of knot diagrams). The \emph{visual knot recognition problem} is to compute a map
\[
\Phi:\ \mathcal{I}\to\mathcal{K},
\]
or approximate such a map, e.g.\ by using a NN. Recognition must be invariant under planar isotopies and the Reidemeister moves, because these preserve the knot type. In particular, a classifier that relies on local textures or incidental background features will fail to generalize; the discriminative signal is encoded by the combinatorics of strands and over/under information at crossings.

Several features distinguish this from standard object recognition (``cat vs.\ dog''): 
\begin{enumerate}[resume]

\item The label is a topological class, invariant under continuous deformations of the rope and changes of camera pose (projective and photometric nuisance variables);

\item The informative content is concentrated in a one-dimensional filament with many self-occlusions (since we are looking at a projection of a 3d object);

\item A cat and a dog are quite different, but knots often look very alike, cf.\ \autoref{fig:pair2}.

\item A cat and a dog differ locally, but no such local distinction can be made for knots as two knot diagrams of the same knot can be wildly different, cf.\ \autoref{Fig:Culprit}.

\end{enumerate}
Thus, the task couples low-level vision (segmentation and centerline extraction) with global symbolic disambiguation (over/under at crossings) under strong consistency constraints, and as we recalled above in \autoref{SS:MathDetection}, the problem itself is very difficult.

\begin{Remark}
Our discussion is restricted to honest knot diagrams and deliberately ignores practical complications. 
In real images, ambiguities are unavoidable: poor picture quality, camera angle, lighting, or physical 
contact can obscure which strand passes over, or worse, introduce classic line-drawing interpretation 
pitfalls. For instance, here is a deliberately poor picture of the knot from \autoref{Eq:Knot}:
\begin{gather*}
\includegraphics[height=4cm]{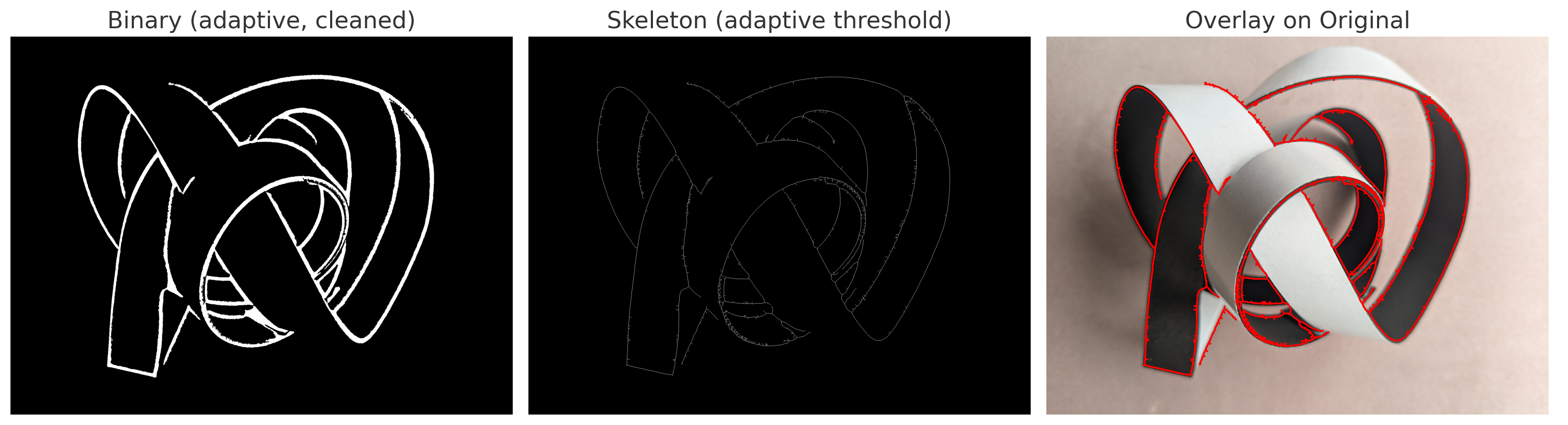}
\end{gather*}
The problem is that, even after skeletonization and cleaning, the result does not resemble a usable knot diagram. 
Any realistic pipeline must therefore disentangle and subsequently reconcile metric appearance from underlying topological structure.
\end{Remark}


\subsection{A subproblem: knot diagram recognition.}

Similarly as above, let $\mathcal{D}$ be the set of knot diagrams. The \emph{visual knot diagram recognition problem} is to compute a map
\[
\Psi:\ \mathcal{I}\to\mathcal{D},
\]
or approximate such a map, e.g.\ by using a NN. 

Here we want to represent a diagram by a PD presentation. Recall that a PD presentation of a knot diagram labels all of its edges, when seen as a planar graph, with non-repeating numbers $\{1,\dots,r\}$, where $r$ is the number of edges. Each edge is then adjacent to two crossings, which induces a labeling of the crossings. We remember the crossings as symbols $X[i,j,k,l]$, where $i$, $j$, $k$ and $l$ are the labels of the edges around that crossing, starting from the incoming lower edge (the one with the smaller label) and proceeding counterclockwise. For example:
\begin{gather*}
PD[{\color{spinach}X[1, 5, 2, 4]},{\color{orchid}X[3, 1, 4, 6]},{\color{tomato}X[5, 3, 6, 2]}]
\leftrightsquigarrow
\begin{tikzpicture}[anchorbase]
\node at (0,0) {\includegraphics[width=0.2\textwidth]{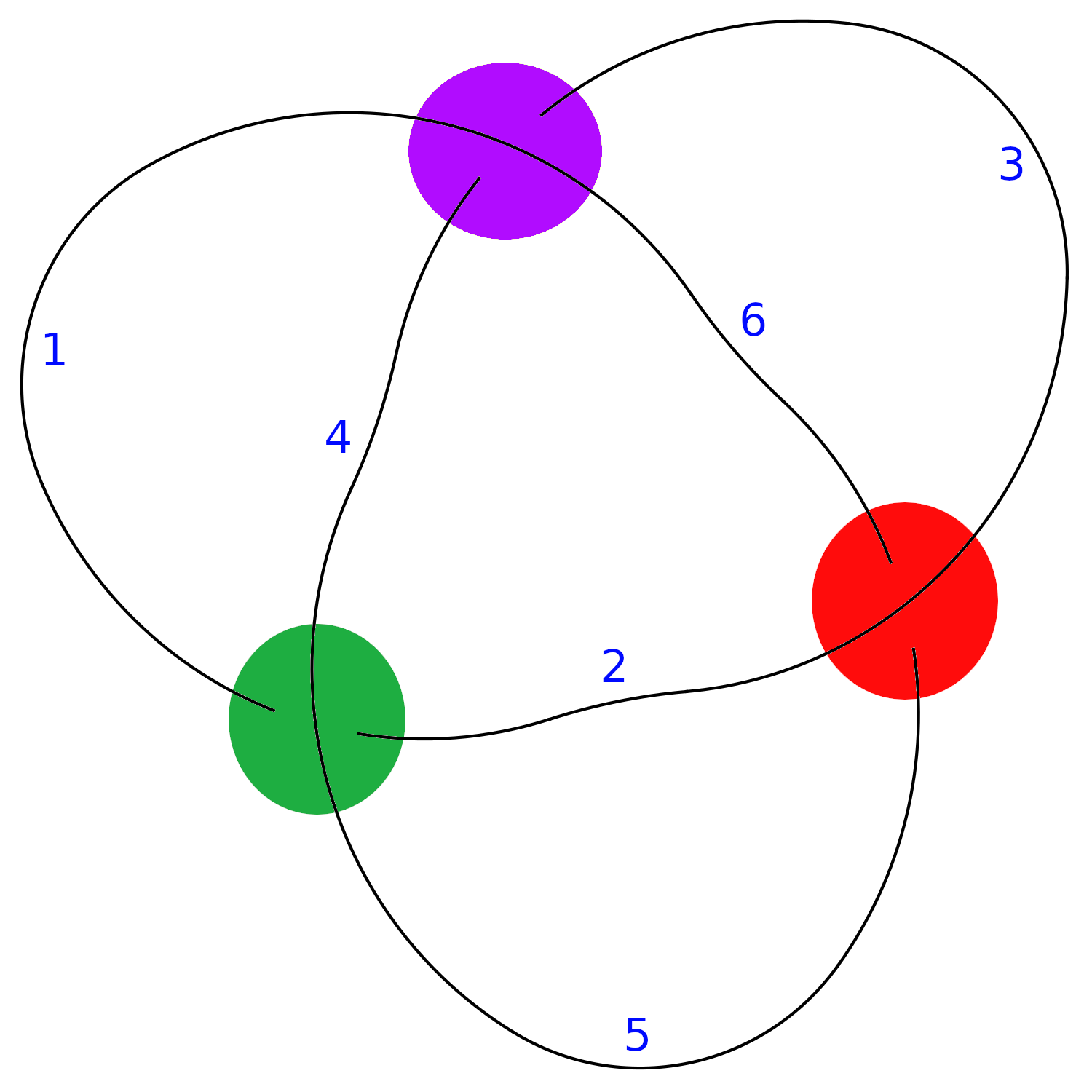}};
\end{tikzpicture}
,
\\
PD[X[4, 2, 5, 1], X[8, 6, 1, 5], X[6, 3, 7, 4], X[2, 7, 3, 8]]
\leftrightsquigarrow
\begin{tikzpicture}[anchorbase]
\node at (0,0) {\includegraphics[width=0.2\textwidth]{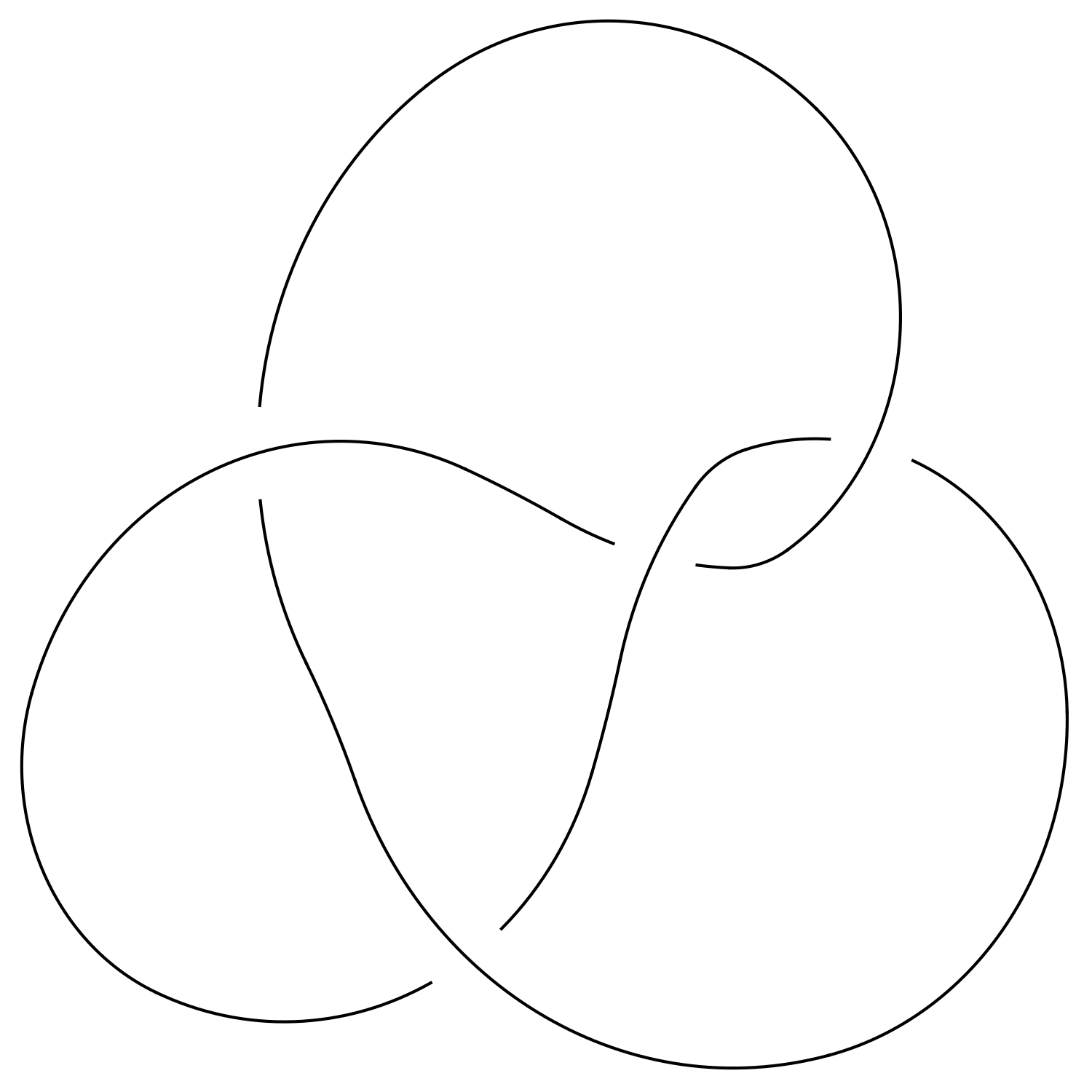}};
\end{tikzpicture}
.
\end{gather*}
(We only added the labels and highlighted the crossings for the trefoil knot at the top.) Of course, we could also just remember e.g. the trefoil as the tensor $[[1, 5, 2, 4],[3, 1, 4, 6],[5, 3, 6, 2]]$, but for clarity one often writes the above.

Coming back to the computation of the map $\Psi$,
a conceptually clean approach is a two-stage ``perception $\longmapsto$ diagram'' factorization in contrast to direct end-to-end classification of raw images into knot types:
\begin{enumerate}[resume]

\item \textit{Rope perception.} Detect and segment the rope (instance-level if multiple ropes are present), then extract a one-pixel centerline (skeleton) or similar. For example:
\begin{gather*}
	\begin{tikzpicture}[anchorbase]
		\node at (0,0) {\reflectbox{\includegraphics[height=4cm]{figs/knotpicture}}};
	\end{tikzpicture}
	\mapsto
	\begin{tikzpicture}[anchorbase]
		\node at (0,0) {\reflectbox{\includegraphics[height=4cm]{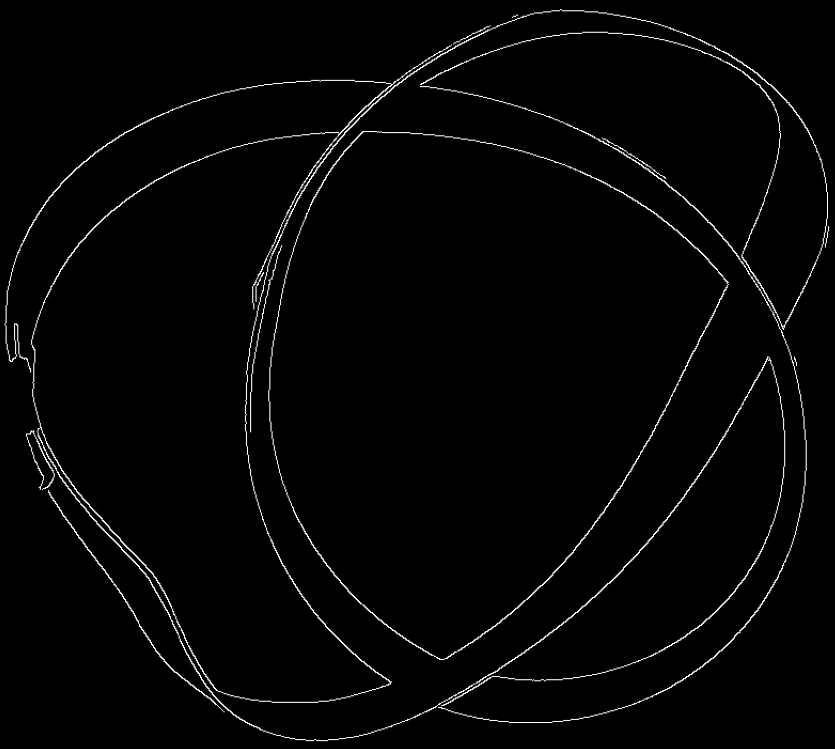}}};
	\end{tikzpicture}
\end{gather*}
This step can be posed with modern DLO (deformable linear object) segmentation and tracing modules, together with topology-preserving skeletonization.

\item \textit{Diagram reconstruction.} From the centerline, detect candidate crossings (degree-4 junctions), classify each as over/under, and produce a planar 4-regular graph with crossing signs; from this, compute a combinatorial code (e.g.\ PD presentations) representing a knot diagram. Because reconstruction must be globally consistent, local over/under votes should be regularized by a global constraint (e.g.\ planarity and 2-in/2-out flow at every vertex), and ambiguities should be resolved using multiple frames or controlled viewpoints when available.

\end{enumerate}
This factorization has three advantages. First, it exposes the source of hardness: classification errors at even a single over/under decision can flip the knot type. 
For example, there is only a small difference between the following two pictures, but the knot types are wildly different:
\begin{gather*}
	\begin{tikzpicture}[anchorbase]
		\node at (0,0) {\reflectbox{\includegraphics[height=4cm]{figs/knotpicture2}}};
	\end{tikzpicture}
	\text{ versus }
	\begin{tikzpicture}[anchorbase]
		\node at (0,0) {\reflectbox{\includegraphics[height=4cm]{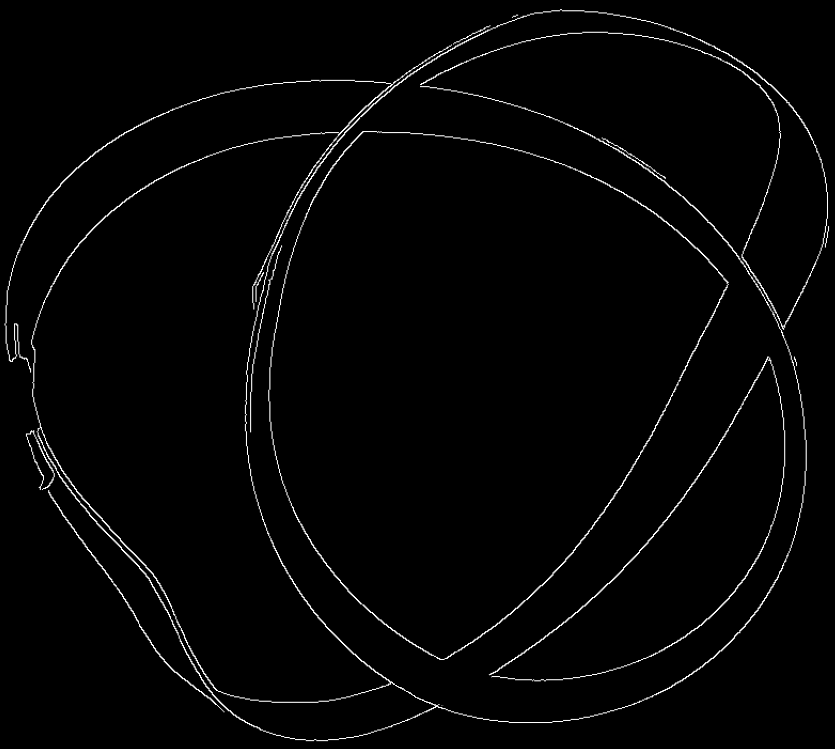}}};
	\end{tikzpicture}
\end{gather*}
Second, it yields interpretable intermediate outputs for debugging (masks, centerlines, junction labels, and a PD presentation). Third, it allows training objectives that couple pixel losses with structural losses (e.g.\ penalties for non-4-regular junctions or forbidden junction labelings).

Why this is a strong benchmark?
Compared to standard recognition tasks, knot diagram recognition stresses capabilities that modern vision systems often lack:
\begin{enumerate}[resume]
	
\item \textit{Topological invariance.} The desired label is invariant under a large group of nuisance transformations such as planar isotopies. Models must learn to ignore photometric and geometric changes that preserve the underlying topology.
\item \textit{Self-occlusion reasoning.} Correctly identifying which strand is on top at many crossings requires long-range, globally consistent reasoning (``explaining away'' shadows, specularities, and contact).
\item \textit{Symbolic decoding.} The output is discrete but determined by a combinatorial object extracted from an image; errors are naturally structured (e.g.\ a small set of wrongly classified crossings).
\item \textit{Combinatorial scaling.} Difficulty scales with crossing number; controlled benchmarks can therefore test generalization along a clear complexity axis.

\end{enumerate}
\begin{gather*}
	\fcolorbox{orchid!50}{spinach!10}{\mystrut``We expect this to be doable; an easier subproblem is discussed in \autoref{S:CrossingDetection}.''} 
\end{gather*}

\section{Our strategy}\label{S:Strategy}

We now explain how we intend to tackle the knot detection problem via image recognition.

\subsection{Reminder: the quantum knot invariants}

We work over $\C(q)$ for a formal variable $q$ that potentially also has roots like $q^{1/k}$.
Following \cite{TuZh-knot-data}, we define a \emph{quantum invariant} $Q(\mathfrak{g},V,\epsilon)$
as a knot/link invariant constructed, as, for example, in \cite{ReTu-invariants-3-manifolds-qgroups,We-knot-invariants}, by selecting a complex semisimple Lie algebra $\mathfrak{g}$, or a similar object, and a simple complex representation $V$ of $\mathfrak{g}$. We let $\epsilon=0$ denote the uncategorified version and $\epsilon=1$ represents its categorified counterpart. A key example is the Jones polynomial, given by $(\mathfrak{sl}_{2},\C^{2},0)$, where $\C^{2}$ is the vector representation of $\mathfrak{sl}_{2}$.

The uncategorified quantum invariants $Q=Q_{\mathfrak{g},V,0}$ can be defined and computed as follows. For a knot, fix a Morse presentation of the knot, arranged vertically.

\begin{Remark}
Given a PD presentation with $n$ crossings, one can algorithmically
produce a Morse presentation with $O(n)$ events in near-linear time.
This follows from \cite{MR1780500}. We henceforth will ignore the difference
between a PD presentation and a Morse presentation. (The caveat to keep in mind is that this does not produce the optimal Morse presentation, i.e.\ with the minimal number of strings, just some Morse presentation.)
\end{Remark} 

For simplicity, assume that $V$ is self-dual. In the Morse presentation, we have four basic pieces and an identity, that we name as
\begin{gather*}
R=
\begin{tikzpicture}[anchorbase,scale=1]
\draw[usual] (0.5,0) to[out=90,in=270] (0,0.5);
\draw[usual,crossline] (0,0) to[out=90,in=270] (0.5,0.5);
\end{tikzpicture}
,\quad
R^{-1}=
\begin{tikzpicture}[anchorbase,scale=1]
\draw[usual] (0,0) to[out=90,in=270] (0.5,0.5);
\draw[usual,crossline] (0.5,0) to[out=90,in=270] (0,0.5);
\end{tikzpicture}
,\quad
cap=
\begin{tikzpicture}[anchorbase,scale=1]
\draw[usual] (0,0) to[out=90,in=180] (0.25,0.25) to[out=0,in=90] (0.5,0);
\end{tikzpicture}
,\quad
cup=
\begin{tikzpicture}[anchorbase,scale=1]
\draw[usual] (0.5,0) to[out=270,in=0] (0.25,-0.25) to[out=180,in=270] (0,0);
\end{tikzpicture}
,\quad
id=
\begin{tikzpicture}[anchorbase,scale=1]
\draw[usual] (0.5,0) to[out=90,in=270] (0.5,0.5);
\end{tikzpicture}
\ .
\end{gather*}
We associate these to linear maps (matrices upon choice of basis) denoted with the same symbols
\begin{gather}\label{Eq:maps}
R,R^{-1}\colon V_{q}\otimes V_{q}\to V_{q}\otimes V_{q}
,\quad
cap\colon V_{q}\otimes V_{q}\to\C(q)
,\quad
cup\colon \C(q)\to V_{q}\otimes V_{q}
,\quad
id\colon V_{q}\to V_{q},v\mapsto v
.
\end{gather}
Here, $V_{q}$ is a representation of a quantum group associated with $\mathfrak{g}$ that dequantizes to the $\mathfrak{g}$ representation $V$, and all the above maps are intertwiners for the quantum group that dequantize to, respective to above, the flip maps, the pairing and coparing of $\mathfrak{g}$ representations, and the identity.

Horizontal composition is then the tensor product (Kronecker product upon choice of basis).
Write $id\otimes\cdots\otimes id$ with $k$ factors simply as $k$.
Then for the figure eight knot this is could be
\begin{gather*}
\begin{tikzpicture}[anchorbase]
	\node at (0,0) {\includegraphics[height=4cm]{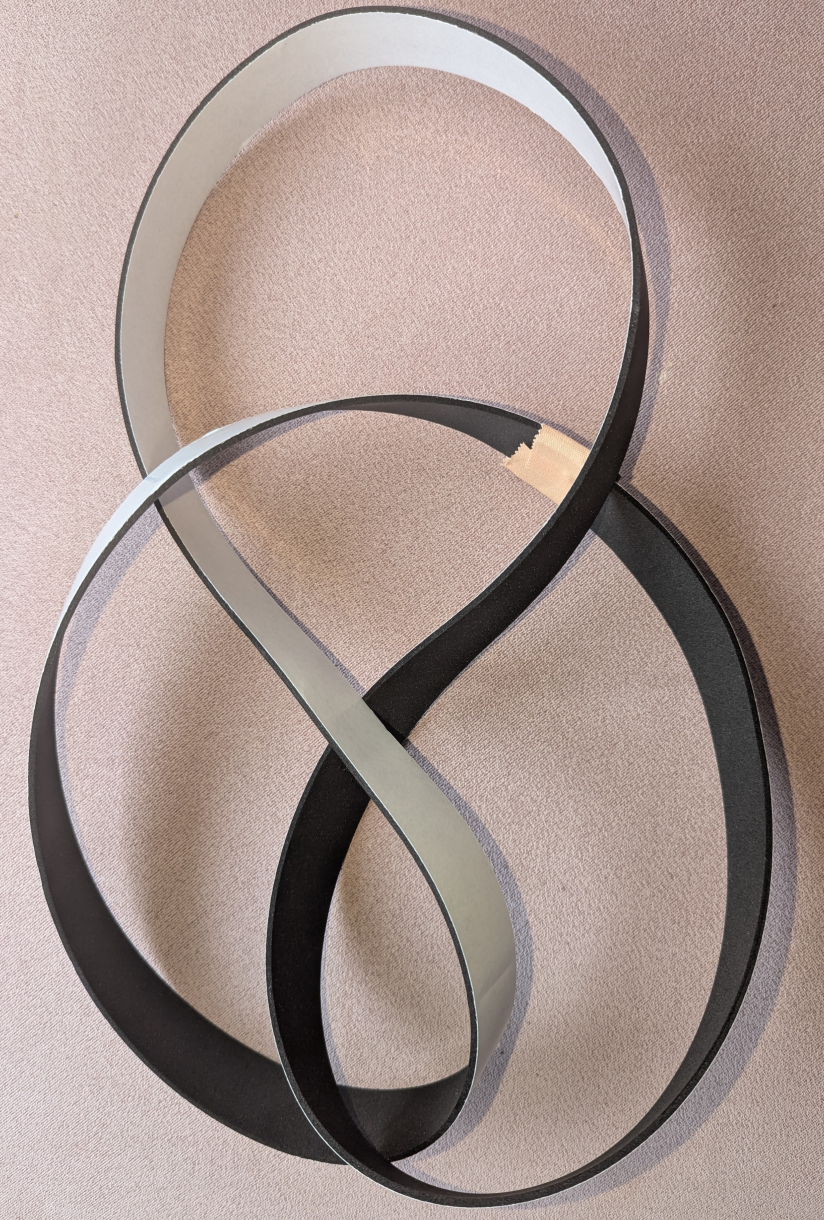}};
\end{tikzpicture}
\mapsto
\begin{tikzpicture}[anchorbase,scale=1]
\draw[usual] (0.5,0) to[out=90,in=270] (0,0.5);
\draw[usual,crossline] (0,0) to[out=90,in=270] (0.5,0.5);
\draw[usual] (1,0) to[out=90,in=270] (1,0.5);
\draw[usual] (0,0.5) to[out=90,in=270] (0,1);
\draw[usual] (0.5,0.5) to[out=90,in=270] (1,1);
\draw[usual,crossline] (1,0.5) to[out=90,in=270] (0.5,1);
\draw[usual] (0.5,1) to[out=90,in=270] (0,1.5);
\draw[usual,crossline] (0,1) to[out=90,in=270] (0.5,1.5);
\draw[usual] (1,1) to[out=90,in=270] (1,1.5);
\draw[usual] (0,1.5) to[out=90,in=270] (0,2);
\draw[usual] (0.5,1.5) to[out=90,in=270] (1,2);
\draw[usual,crossline] (1,1.5) to[out=90,in=270] (0.5,2);
\draw[usual] (1,2) to[out=90,in=180] (1.25,2.25) to[out=0,in=90] (1.5,2) to (1.5,0) to[out=270,in=0] (1.25,-0.25) to[out=180,in=270] (1,0);
\draw[usual] (0.5,2) to[out=90,in=180] (1.25,2.5) to[out=0,in=90] (2,2) to (2,0) to[out=270,in=0] (1.25,-0.5) to[out=180,in=270] (0.5,0);
\draw[usual] (0,2) to[out=90,in=180] (1.25,2.75) to[out=0,in=90] (2.5,2) to (2.5,0) to[out=270,in=0] (1.25,-0.75) to[out=180,in=270] (0,0);
\end{tikzpicture}
\mapsto
\scalebox{0.65}{$\begin{tikzcd}[ampersand replacement=\&,column sep=0.15em,row sep=0.5em]
0\otimes cap \otimes 0
\ar[d,-,"\circ"]
\\
1\otimes cap\otimes 1
\ar[d,-,"\circ"]
\\
2\otimes cap\otimes 2
\ar[d,-,"\circ"]
\\
1\otimes R^{-1}\otimes 3
\ar[d,-,"\circ"]
\\
0\otimes R\otimes 4
\ar[d,-,"\circ"]
\\
1\otimes R^{-1}\otimes 3
\ar[d,-,"\circ"]
\\
0\otimes R\otimes 4
\ar[d,-,"\circ"]
\\
2\otimes cup\otimes 2
\ar[d,-,"\circ"]
\\
1\otimes cup\otimes 1
\ar[d,-,"\circ"]
\\
0\otimes cup \otimes 0
\end{tikzcd}$}
,
\end{gather*}
with composition, for example from bottom to top.

The maps in \autoref{Eq:maps} are not uniquely determined from what we wrote above. There are some choices involved, but they are mostly irrelevant and just rescale the quantum invariant. To be completely explicit, we follow the conventions used in \cite{KnotTheory}. For example, in coordinates,
\begin{gather*}
R=
\begin{pmatrix}
q^{1/2} & 0 & 0 & 0
\\
0 & 0 & q & 0
\\
0 & q & q^{1/2}-q^{3/2} & 0
\\
0 & 0 & 0 & q^{1/2}
\end{pmatrix}
,
\end{gather*}
is the choice of $R$-matrix for the Jones polynomial.

To analyze its computational complexity, let $N=\dim_{\C}V$, and let $p(n)$ denote some polynomial in $n$. With a bit of care, see \cite[Theorem 1.1]{Ma-complexity-quantum}, one can show that
\begin{gather*}
Q_{\mathfrak{g},V,0}\in O(p(n)N^{3\sqrt{n}/2}).
\end{gather*}
Thus, computing quantum invariants using $R$-matrices is superpolynomial but subexponential in $n$, with the leading factor determined by
the dimension of the underlying representation.

\begin{Remark}
This is not surprising if one keeps in mind that the main difficulty in this computational approach is not the number of crossings, but rather the number of strands, since this corresponds to tensoring $V$, which, even in decompositions \cite{CoOsTu-growth,KhSiTu-monoidal-cryptography}, behaves exponentially in $N$.
\end{Remark}

Computing the categorified quantum invariants is less homogeneous and more complicated, and we will not recall how this works. The currently known algorithms 
(roughly) show
\begin{gather*}
Q_{\mathfrak{g},V,1}\in_{\approx} O(N^{n}).
\end{gather*}
Here $\in_{\approx}$ means ``roughly expected'' (we do not know any formal statement).
In any case, we will argue below that one probably doesn't want to use them for the task at hand.

The ones we mostly use below from this infinite zoo are the \emph{Jones polynomial} $(\mathfrak{sl}_{2},\C^{2},0)$ (for the vector representation), and its categorification \emph{Khovanov homology} $(\mathfrak{sl}_{2},\C^{2},1)$. By the above:
\begin{gather*}
\text{Jones}\in_{\approx} O(2^{\sqrt{n}}),\quad
\text{Khovanov}\in_{\approx} O(2^{n}).
\end{gather*}
For the type of knot diagrams we have in mind with $\leq 30$ crossings, the calculation of the invariants is therefore essentially instant.

\subsection{How good are quantum knot invariants?}

Quantum invariants are traditionally regarded as tools for distinguishing knots.
However, motivated by a form of ``folk wisdom'', and supported by prior theoretical \cite{St-number-polynomials,TuZh-knot-data,LaTuVaZh-big-data-2} and empirical work \cite{LeHaSa-jones,DlGuSa-mapper,LaTuVa-big-data,DlGuSa-data,TuZh-knot-data,LaTuVaZh-big-data-2}, it is expected that many (or even most) quantum invariants fail to detect most knots in a probabilistic sense.

We now summarize the main findings of \cite{TuZh-knot-data,LaTuVaZh-big-data-2} regarding the question: ``How effective are quantum invariants as tools for distinguishing knots?''
We start with a theoretical result, formulated only for the Jones polynomial but true in more generality:

\begin{Theorem}\label{T:Main}
The Jones polynomial detects alternating links with probability zero, and the detection probability decays exponentially with the crossing number.
\end{Theorem}

\begin{proof}
This is \cite[Theorem 3.5]{TuZh-knot-data} or \cite{LaTuVaZh-big-data-2}. We only stress here that the proof given therein could potentially show that many (most or even all?) quantum invariants suffer from the same decay.
\end{proof}

For experimental data we follow \cite{TuZh-knot-data}. The question is how many distinct values quantum invariants take on our list of knots, and we measure this as a percentage. In other words, we want to know
\begin{gather*}
Q(n)^{\%}=
\#\{Q(K) \mid K\in\mathcal{K}_{n}\}/
\#\mathcal{K}_{n}.
\end{gather*}
These are the \emph{distinct values} $Q$ takes.

\begin{figure}[!ht]
\begin{center}
\includegraphics[height=6cm]{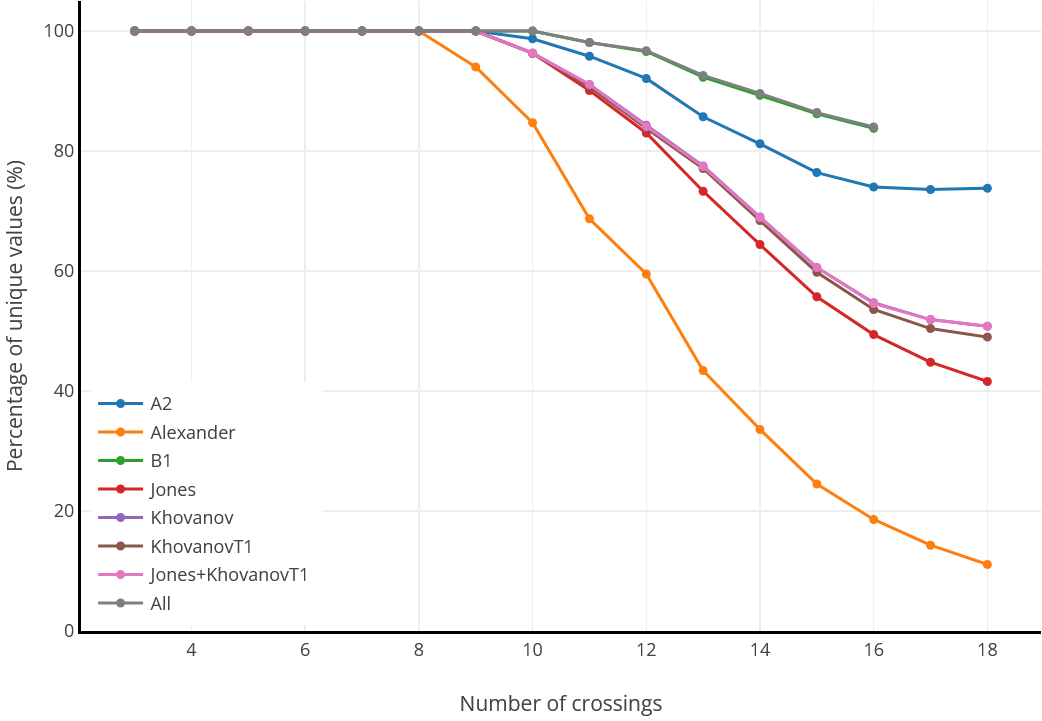}
\end{center}
\caption{Percentage of unique values.}
\label{fig:percent-uniq-val}
\end{figure}

Recall that A, J, and K stand for Alexander, Jones, and Khovanov, respectively. Let ``All''
mean that we take all of them together, and we use J+KT1 to take Jones and KT1 together. 
The data is displayed in \autoref{fig:percent-uniq-val} and the precise values are listed in \autoref{table:percent-uniq-val}.

\begin{table}[H]
  \centering
  \begingroup
  \setlength{\fboxsep}{0pt}
  \renewcommand{\arraystretch}{1.05}
  \setlength{\tabcolsep}{4pt}
  \fcolorbox{tomato!50}{white}{%
    \rowcolors{2}{spinach!25}{}
    \begin{tabular}{
      c
      S[table-format=3.1]
      S[table-format=3.1]
      S[table-format=3.1]
      S[table-format=3.1]
      S[table-format=3.1]
      S[table-format=3.1]
      S[table-format=3.1]
      S[table-format=3.1]
    }
      \rowcolor{orchid!50}
      {$n$} & {A2} & {A} & {B1} & {J} & {K} & {KT1} & {J+KT1} & {All} \\
      \midrule
       3  & 100.0 & 100.0 & 100.0 & 100.0 & 100.0 & 100.0 & 100.0 & 100.0 \\
       4  & 100.0 & 100.0 & 100.0 & 100.0 & 100.0 & 100.0 & 100.0 & 100.0 \\
       5  & 100.0 & 100.0 & 100.0 & 100.0 & 100.0 & 100.0 & 100.0 & 100.0 \\
       6  & 100.0 & 100.0 & 100.0 & 100.0 & 100.0 & 100.0 & 100.0 & 100.0 \\
       7  & 100.0 & 100.0 & 100.0 & 100.0 & 100.0 & 100.0 & 100.0 & 100.0 \\
       8  & 100.0 & 100.0 & 100.0 & 100.0 & 100.0 & 100.0 & 100.0 & 100.0 \\
       9  & 100.0 &  94.0 & 100.0 & 100.0 & 100.0 & 100.0 & 100.0 & 100.0 \\
      10  &  98.7 &  84.7 & 100.0 &  96.3 &  96.3 &  96.3 &  96.3 & 100.0 \\
      11  &  95.8 &  68.7 &  98.1 &  90.1 &  91.1 &  90.7 &  91.1 &  98.1 \\
      12  &  92.1 &  59.5 &  96.6 &  83.0 &  84.3 &  83.8 &  84.1 &  96.7 \\
      13  &  85.7 &  43.4 &  92.3 &  73.3 &  77.5 &  77.1 &  77.4 &  92.6 \\
      14  &  81.2 &  33.6 &  89.3 &  64.4 &  69.0 &  68.4 &  68.9 &  89.6 \\
      15  &  76.4 &  24.5 &  86.2 &  55.7 &  60.6 &  59.8 &  60.6 &  86.4 \\
      16  &  74.0 &  18.6 &  83.8 &  49.4 &  54.7 &  53.6 &  54.6 &  84.0 \\
      17  &  73.6 &  14.3 &  \text{$\approx82?$} &  44.8 &  51.9 &  50.4 &  51.9 &  \text{$\approx82?$} \\
      18  &  73.8 &  11.1 &  \text{$\approx80?$} &  41.6 &  50.8 &  49.0 &  50.8 &  \text{$\approx80?$} \\
    \end{tabular}%
  }
  \endgroup
  \caption{Percentages of unique values; copyable data. The data for the B1 invariant only goes up to 16 crossings, and we made a guess for 17 and 18 crossings.}
  \label{table:percent-uniq-val}
\end{table}

\begin{figure}[ht]
\centering
\includegraphics[width=0.6\textwidth]{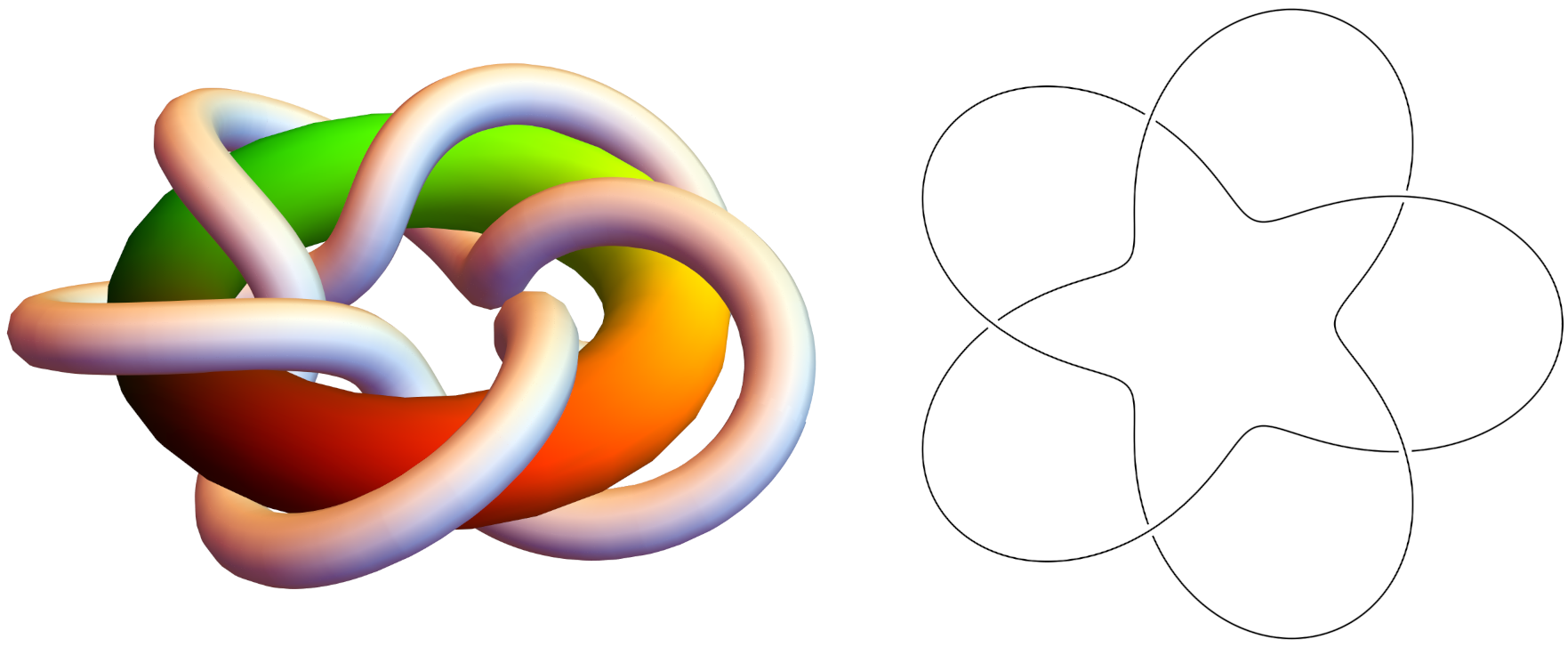}
\caption{A torus knot \(T(p,q)\), with \(\gcd(p,q)=1\), lies on the surface of a (standardly embedded) torus, winding \(p\) times around a meridian and \(q\) times around a longitude; shown is \(T(2,5)\) and its diagram. These knots have very characteristic diagrams.
}
\label{Fig:Torus}
\end{figure}

\begin{figure}[ht]
\centering
\includegraphics[width=0.6\textwidth]{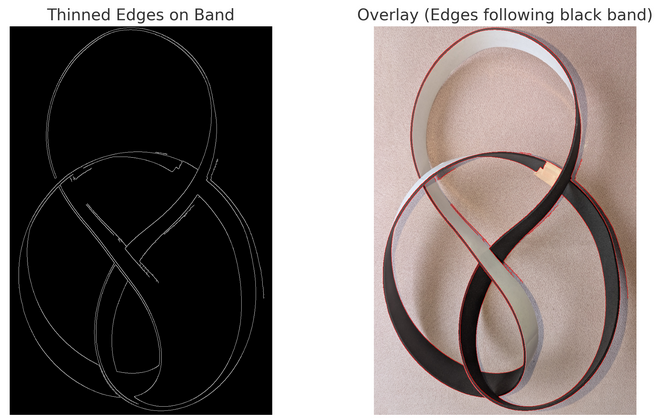}
\caption{A twist knot is obtained by repeatedly twisting a closed loop and then linking the ends together, here the figure eight knot $4_1$ with two twists. Creating this knot and a reasonably nice picture of it from a 2cm width and 150cm length strand was nontrivial.
}
\label{Fig:Twist}
\end{figure}

\cite{LaTuVaZh-big-data-2} gives additional data for many categorified quantum invariants, showing the same trend.
In other words, some (potentially most or even all) quantum invariants fail to distinguish the vast majority of knots and links.
However, and that is a key feature, cf. \autoref{table:percent-uniq-val}:
\begin{gather*}
\fcolorbox{orchid!50}{spinach!10}{\mystrut``Quantum invariants are unreasonably good for small (or easy) knots!''} 
\end{gather*}
Here and throughout, let us say a knot is small if it has a diagram with few crossings.
Moreover, easy knots are
those for which a small set of standard invariants suffices to
distinguish them from all others. Typical examples include torus knots, cf. \autoref{Fig:Torus}, which are
known to be characterized by quantum (and a bit beyond) invariants
\cite{MR3604373}. Another fundamental family of easy knots are the twist knots, see \autoref{Fig:Twist}, such as $5_2$, and they are usually easy to detect, see e.g. \cite[Theorem 1.6]{BaldwinSivekNonfibered}.

Recall that prime knots are commonly indexed in the Alexander--Briggs notation \(K=A_B\), where \(A\) is the minimal crossing number of all diagrams of $K$ and \(B\) enumerates the (prime) knots with \(A\) crossings; see \autoref{fig:DNA1}. For later reference, the first few torus and twist knots (in this notation) are the unknot and $3_1,5_1,7_1,8_{19}$ (torus), and 
$3_1,4_1,5_2,6_1,7_2,8_1$ (twist), and no other knots with $A\leq 8$.

\subsection{Distribution of knots in pictures}

\begin{figure}[ht]
\centering
\includegraphics[width=0.9\textwidth]{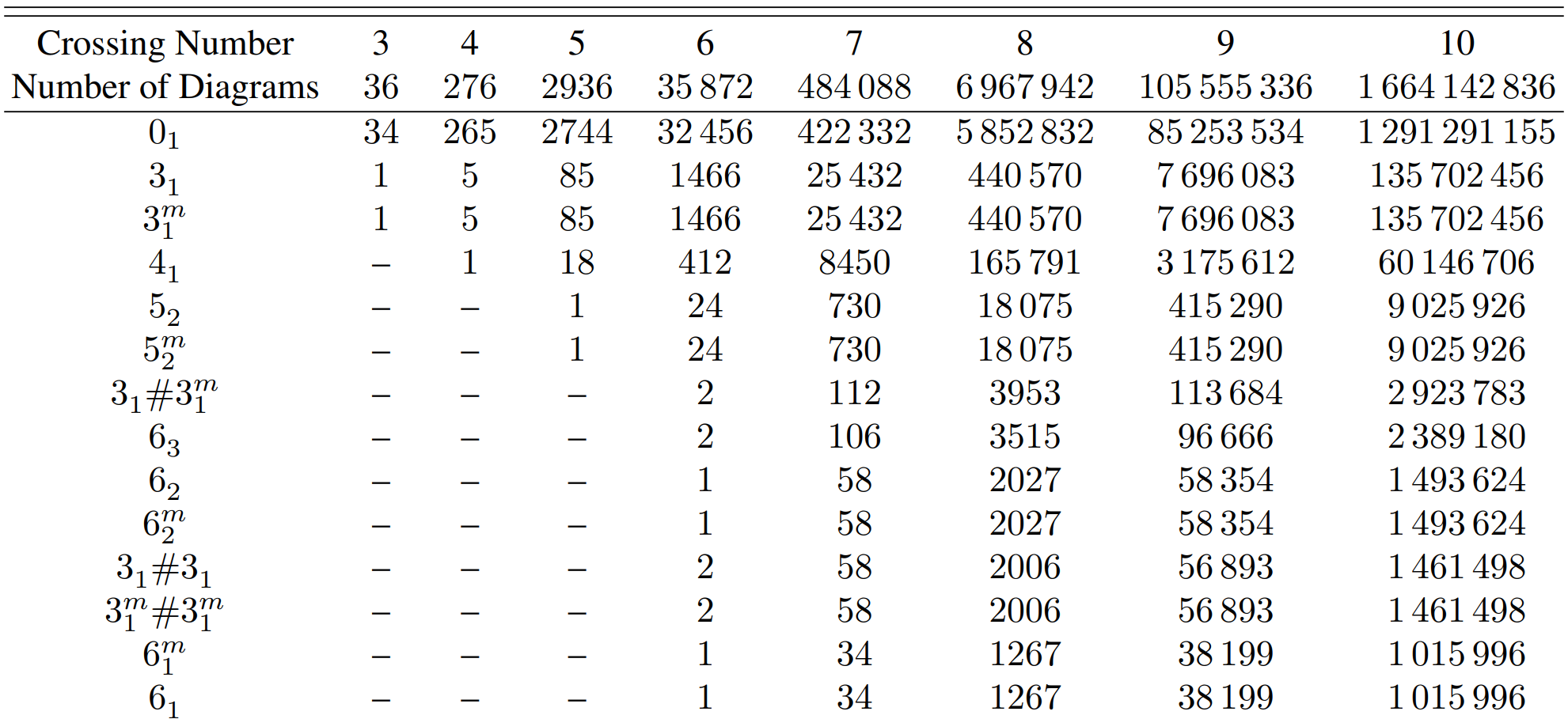}
\caption{The percentage of different knots among 10 crossing diagrams in a (certain) random model. For example, $\approx$77\% of all ten crossing diagrams are unknots.
\\
{\tiny Picture from \cite{MR3556174}.}}
\label{Fig:Random}
\end{figure}

\begin{figure}[ht]
\centering
\includegraphics[width=0.7\textwidth]{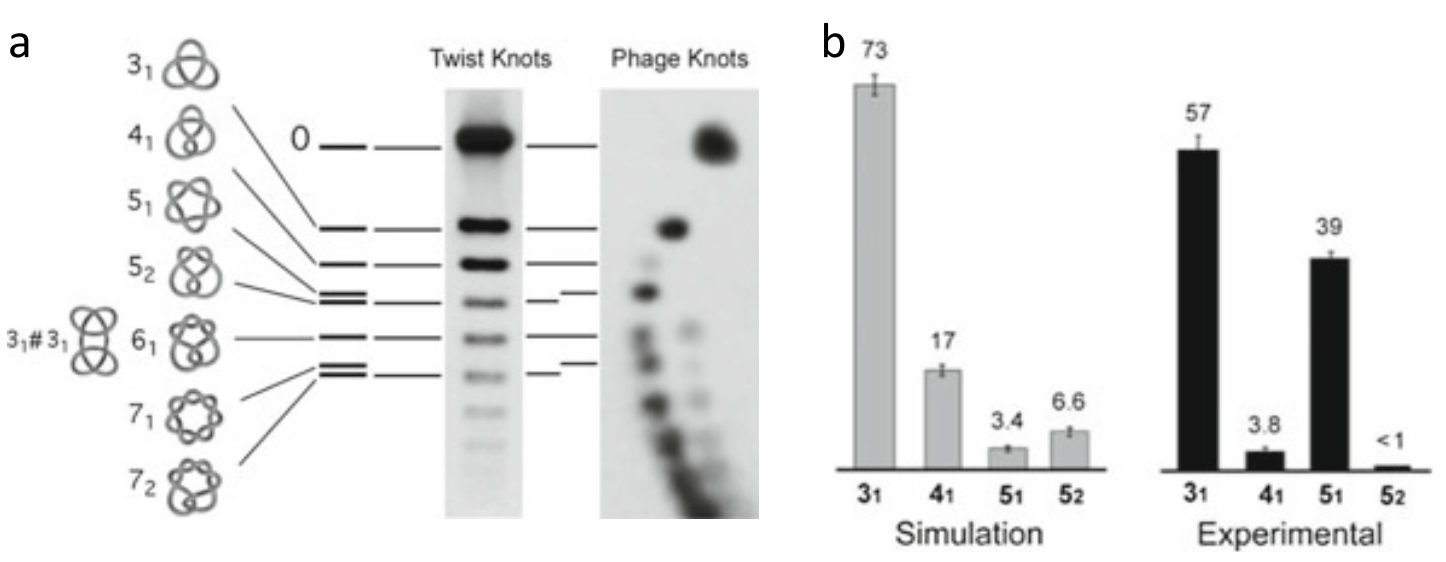}
\caption{As in \autoref{Fig:Random}, but for DNA knots. Note that torus knots are overrepresented.
\\
{\tiny Picture from \cite{DNAchiral2005,Micheletti2022}.}}
\label{Fig:DNAreal}
\end{figure}

The crucial observation is:
\begin{gather*}
\fcolorbox{orchid!50}{spinach!10}{\mystrut``Almost all knot diagrams correspond to small (or easy) knots!''} 
\end{gather*}
This is true in random models, cf.\ \autoref{Fig:Random}, as well as real life, cf. \autoref{Fig:DNAreal}. We will now argue why we expect this in pictures of knots.

\begin{figure}[ht]
\centering
\includegraphics[width=0.7\textwidth]{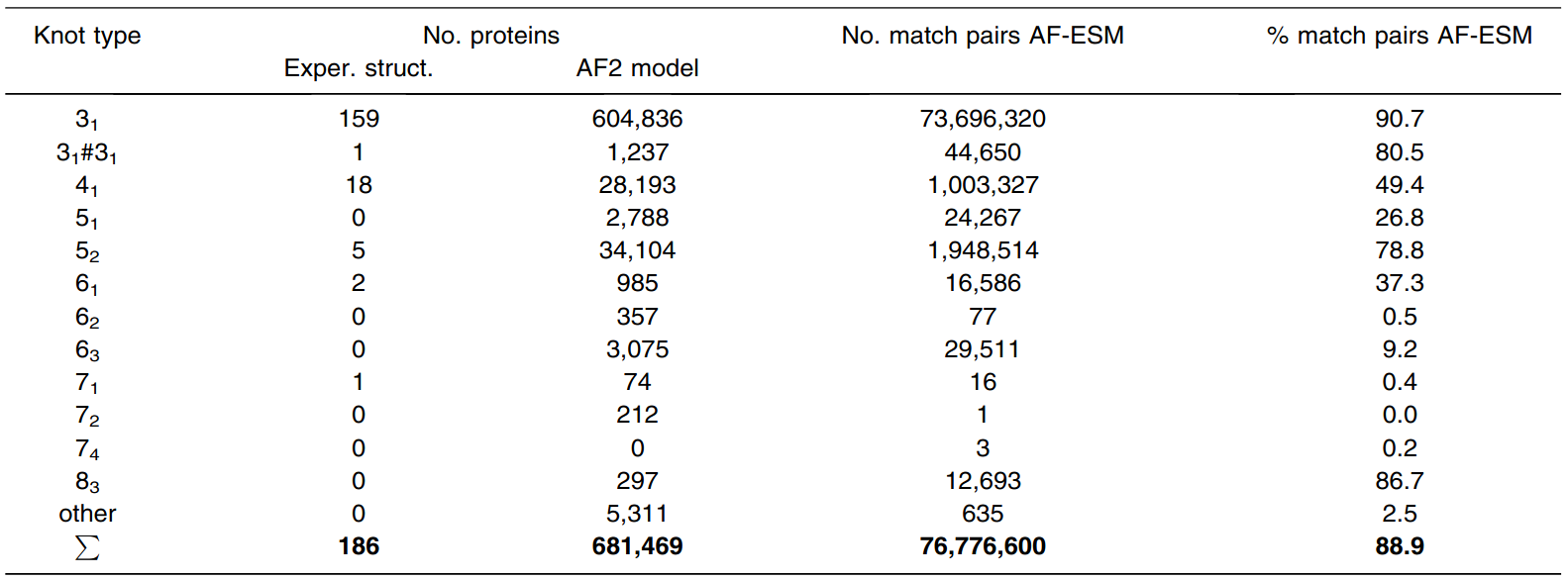}
\caption{As in \autoref{Fig:Random}, but for protein knots. Note that twist knots are overrepresented.
\\
{\tiny Picture from \cite{AlphaFoldKnots2024}.}}
\label{Fig:Proteinreal}
\end{figure}

To get started, let us explain \autoref{Fig:Random}; details can be found in \cite{MR3556174}. Say a \emph{random knot} is obtained by randomly drawing a simple closed curve and then randomly selecting crossing information. For example:
\begin{gather*}
\begin{tikzpicture}[anchorbase,scale=1]
	\draw[usual] (1,0) to[out=90,in=270] (0.5,0.5);
	\draw[usual] (0.5,0) to[out=90,in=270] (1,0.5);
	\draw[usual] (1,0.5) to[out=90,in=270] (0.5,1);
	\draw[usual] (0.5,0.5) to[out=90,in=270] (1,1);
	\draw[usual] (1,1) to[out=90,in=270] (0.5,1.5);
	\draw[usual] (0.5,1) to[out=90,in=270] (1,1.5);
	\draw[usual] (1,1.5) to[out=90,in=180] (1.25,1.75) to[out=0,in=90] (1.5,1.5) to (1.5,0) to[out=270,in=0] (1.25,-0.25) to[out=180,in=270] (1,0);
	\draw[usual] (0.5,1.5) to[out=90,in=180] (1.25,2) to[out=0,in=90] (2,1.5) to (2,0) to[out=270,in=0] (1.25,-0.5) to[out=180,in=270] (0.5,0);
\end{tikzpicture}
\rightsquigarrow
\begin{tikzpicture}[anchorbase,scale=1]
	\draw[usual] (1,0) to[out=90,in=270] (0.5,0.5);
	\draw[usual,crossline] (0.5,0) to[out=90,in=270] (1,0.5);
	\draw[usual] (1,0.5) to[out=90,in=270] (0.5,1);
	\draw[usual,crossline] (0.5,0.5) to[out=90,in=270] (1,1);
	\draw[usual] (1,1) to[out=90,in=270] (0.5,1.5);
	\draw[usual,crossline] (0.5,1) to[out=90,in=270] (1,1.5);
	\draw[usual] (1,1.5) to[out=90,in=180] (1.25,1.75) to[out=0,in=90] (1.5,1.5) to (1.5,0) to[out=270,in=0] (1.25,-0.25) to[out=180,in=270] (1,0);
	\draw[usual] (0.5,1.5) to[out=90,in=180] (1.25,2) to[out=0,in=90] (2,1.5) to (2,0) to[out=270,in=0] (1.25,-0.5) to[out=180,in=270] (0.5,0);
\end{tikzpicture}
,
\begin{tikzpicture}[anchorbase,scale=1]
	\draw[usual] (0.5,0) to[out=90,in=270] (1,0.5);
	\draw[usual,crossline] (1,0) to[out=90,in=270] (0.5,0.5);
	\draw[usual] (0.5,0.5) to[out=90,in=270] (1,1);
	\draw[usual,crossline] (1,0.5) to[out=90,in=270] (0.5,1);
	\draw[usual] (0.5,1) to[out=90,in=270] (1,1.5);
	\draw[usual,crossline] (1,1) to[out=90,in=270] (0.5,1.5);
	\draw[usual] (1,1.5) to[out=90,in=180] (1.25,1.75) to[out=0,in=90] (1.5,1.5) to (1.5,0) to[out=270,in=0] (1.25,-0.25) to[out=180,in=270] (1,0);
	\draw[usual] (0.5,1.5) to[out=90,in=180] (1.25,2) to[out=0,in=90] (2,1.5) to (2,0) to[out=270,in=0] (1.25,-0.5) to[out=180,in=270] (0.5,0);
\end{tikzpicture}
;
\text{ or }
\begin{tikzpicture}[anchorbase,scale=1]
	\draw[usual] (0.5,0) to[out=90,in=270] (1,0.5);
	\draw[usual,crossline] (1,0) to[out=90,in=270] (0.5,0.5);
	\draw[usual] (1,0.5) to[out=90,in=270] (0.5,1);
	\draw[usual,crossline] (0.5,0.5) to[out=90,in=270] (1,1);
	\draw[usual] (1,1) to[out=90,in=270] (0.5,1.5);
	\draw[usual,crossline] (0.5,1) to[out=90,in=270] (1,1.5);
	\draw[usual] (1,1.5) to[out=90,in=180] (1.25,1.75) to[out=0,in=90] (1.5,1.5) to (1.5,0) to[out=270,in=0] (1.25,-0.25) to[out=180,in=270] (1,0);
	\draw[usual] (0.5,1.5) to[out=90,in=180] (1.25,2) to[out=0,in=90] (2,1.5) to (2,0) to[out=270,in=0] (1.25,-0.5) to[out=180,in=270] (0.5,0);
\end{tikzpicture}
,
\begin{tikzpicture}[anchorbase,scale=1]
	\draw[usual] (1,0) to[out=90,in=270] (0.5,0.5);
	\draw[usual,crossline] (0.5,0) to[out=90,in=270] (1,0.5);
	\draw[usual] (0.5,0.5) to[out=90,in=270] (1,1);
	\draw[usual,crossline] (1,0.5) to[out=90,in=270] (0.5,1);
	\draw[usual] (1,1) to[out=90,in=270] (0.5,1.5);
	\draw[usual,crossline] (0.5,1) to[out=90,in=270] (1,1.5);
	\draw[usual] (1,1.5) to[out=90,in=180] (1.25,1.75) to[out=0,in=90] (1.5,1.5) to (1.5,0) to[out=270,in=0] (1.25,-0.25) to[out=180,in=270] (1,0);
	\draw[usual] (0.5,1.5) to[out=90,in=180] (1.25,2) to[out=0,in=90] (2,1.5) to (2,0) to[out=270,in=0] (1.25,-0.5) to[out=180,in=270] (0.5,0);
\end{tikzpicture}
,
\begin{tikzpicture}[anchorbase,scale=1]
	\draw[usual] (1,0) to[out=90,in=270] (0.5,0.5);
	\draw[usual,crossline] (0.5,0) to[out=90,in=270] (1,0.5);
	\draw[usual] (1,0.5) to[out=90,in=270] (0.5,1);
	\draw[usual,crossline] (0.5,0.5) to[out=90,in=270] (1,1);
	\draw[usual] (0.5,1) to[out=90,in=270] (1,1.5);
	\draw[usual,crossline] (1,1) to[out=90,in=270] (0.5,1.5);
	\draw[usual] (1,1.5) to[out=90,in=180] (1.25,1.75) to[out=0,in=90] (1.5,1.5) to (1.5,0) to[out=270,in=0] (1.25,-0.25) to[out=180,in=270] (1,0);
	\draw[usual] (0.5,1.5) to[out=90,in=180] (1.25,2) to[out=0,in=90] (2,1.5) to (2,0) to[out=270,in=0] (1.25,-0.5) to[out=180,in=270] (0.5,0);
\end{tikzpicture}
,
\begin{tikzpicture}[anchorbase,scale=1]
	\draw[usual] (0.5,0) to[out=90,in=270] (1,0.5);
	\draw[usual,crossline] (1,0) to[out=90,in=270] (0.5,0.5);
	\draw[usual] (0.5,0.5) to[out=90,in=270] (1,1);
	\draw[usual,crossline] (1,0.5) to[out=90,in=270] (0.5,1);
	\draw[usual] (1,1) to[out=90,in=270] (0.5,1.5);
	\draw[usual,crossline] (0.5,1) to[out=90,in=270] (1,1.5);
	\draw[usual] (1,1.5) to[out=90,in=180] (1.25,1.75) to[out=0,in=90] (1.5,1.5) to (1.5,0) to[out=270,in=0] (1.25,-0.25) to[out=180,in=270] (1,0);
	\draw[usual] (0.5,1.5) to[out=90,in=180] (1.25,2) to[out=0,in=90] (2,1.5) to (2,0) to[out=270,in=0] (1.25,-0.5) to[out=180,in=270] (0.5,0);
\end{tikzpicture}
,
\begin{tikzpicture}[anchorbase,scale=1]
	\draw[usual] (0.5,0) to[out=90,in=270] (1,0.5);
	\draw[usual,crossline] (1,0) to[out=90,in=270] (0.5,0.5);
	\draw[usual] (1,0.5) to[out=90,in=270] (0.5,1);
	\draw[usual,crossline] (0.5,0.5) to[out=90,in=270] (1,1);
	\draw[usual] (0.5,1) to[out=90,in=270] (1,1.5);
	\draw[usual,crossline] (1,1) to[out=90,in=270] (0.5,1.5);
	\draw[usual] (1,1.5) to[out=90,in=180] (1.25,1.75) to[out=0,in=90] (1.5,1.5) to (1.5,0) to[out=270,in=0] (1.25,-0.25) to[out=180,in=270] (1,0);
	\draw[usual] (0.5,1.5) to[out=90,in=180] (1.25,2) to[out=0,in=90] (2,1.5) to (2,0) to[out=270,in=0] (1.25,-0.5) to[out=180,in=270] (0.5,0);
\end{tikzpicture}
,
\begin{tikzpicture}[anchorbase,scale=1]
	\draw[usual] (1,0) to[out=90,in=270] (0.5,0.5);
	\draw[usual,crossline] (0.5,0) to[out=90,in=270] (1,0.5);
	\draw[usual] (0.5,0.5) to[out=90,in=270] (1,1);
	\draw[usual,crossline] (1,0.5) to[out=90,in=270] (0.5,1);
	\draw[usual] (0.5,1) to[out=90,in=270] (1,1.5);
	\draw[usual,crossline] (1,1) to[out=90,in=270] (0.5,1.5);
	\draw[usual] (1,1.5) to[out=90,in=180] (1.25,1.75) to[out=0,in=90] (1.5,1.5) to (1.5,0) to[out=270,in=0] (1.25,-0.25) to[out=180,in=270] (1,0);
	\draw[usual] (0.5,1.5) to[out=90,in=180] (1.25,2) to[out=0,in=90] (2,1.5) to (2,0) to[out=270,in=0] (1.25,-0.5) to[out=180,in=270] (0.5,0);
\end{tikzpicture}
.
\end{gather*}
Note that two of the eight ($=2^3$) possible choices of crossing information give the trefoils (the first two choices above), while the other six choices give unknots (on the right above). So the knot with the lower crossing number (the unknot) is overrepresented in this choice of a three crossing diagram. The same is true in more generality, although the precise distribution depends on the random model, e.g.\ compare 
\autoref{Fig:Random} and \autoref{Fig:DNAreal} and \autoref{Fig:Proteinreal}.

\begin{Remark}
In circular DNA extracted from bacteriophages, the empirical
distribution of knot types is markedly biased toward torus knots, cf. 
\autoref{Fig:DNAreal}.
By contrast, knotted proteins do not show a bias towards torus knots but rather towards twist knots, cf. \autoref{Fig:Proteinreal}.
\end{Remark}

Let us now comment on the expected distribution of knots in pictures.
We write $c(K)$ for the minimal crossing number. We argue that, under natural generative mechanisms for photographs of knotted ropes or cables, most observed knots have small $c(K)$. The claim rests on three pillars:
\begin{enumerate}[resume]

\item Empirical laws from models of random knots;

\item Geometric resource constraints (rope length versus $c(K)$);

\item Selection effects in how pictures are produced.

\end{enumerate}
We now elaborate on the bottom two as the other was addressed above.

\textit{(21).}
Large crossing number cannot be realized economically: for a unit-thickness embedding, the ropelength grows at least on the order of $c(K)^{3/4}$ (and often linearly for broad families), while upper bounds scale at most quadratically, cf.\ \cite{MR1666650}. Thus, to exhibit a knot with large $c(K)$ in a physical rope of fixed thickness one must spend substantial length. In informal terms: with a one-meter rope of a few centimeters diameter, one can reliably tie $3_1$ or $4_1$, but producing $c(K)\gg 10$ without self-contact artifacts is difficult. In fact, even $4_1$ is already difficult to produce, cf. \autoref{Fig:Twist}. Since most casual pictures are made under tight length and field-of-view budgets, the feasible set of types concentrates on small $c(K)$.

\textit{(22).}
Even setting randomness aside, the way pictures are made amplifies small types. Human-tied knots (sailing, climbing, crafts) overwhelmingly use low-complexity patterns; natural occurrences (e.g.\ DNA, proteins, cables) also favor a short list of simple types, see e.g. \autoref{Fig:DNAreal} or \autoref{Fig:Proteinreal}. Datasets, tutorials, and documentation replicate these biases. Hence, the observational distribution over knot types is far from uniform over $\{K:c(K)\le C\}$; it is sharply peaked at small $c(K)$ even when conditioning on ``knotted.''

\begin{Remark}
Counting arguments go the other way: the number of prime knots grows exponentially with $c(K)$ (already for alternating knots) \cite{ErSu-growth-knots}, so uniformly sampling knot types makes ``most'' types large. Our claim concerns observational measures induced by physical, human, and imaging processes, not a uniform prior on types. Likewise, at very large chain lengths the typical $c(K)$ increases; if one photographs extremely long, highly confined strings, complex knots become common, by e.g. \cite{MR951053}.
\end{Remark}

To summarize, for moderate lengths and realistic imaging conditions, (20) random-knot spectra overwhelmingly favor small types; (21) ropelength constraints penalize large $c(K)$; and (22) human and natural data sources select simple patterns. These effects compound, so most knots that show up in pictures are small.

\subsection{Strategy: NN and traditional algorithms}

We now make our strategy precise and justify why it is both (i) mathematically principled and (ii) computationally feasible in the regime of interest (small crossing number, and the ``unreasonably good'' behavior of quantum invariants on that regime).

The picture summarizing the strategy is:
\begin{gather*}
\fcolorbox{orchid!50}{spinach!10}{\mystrut$\begin{tikzpicture}[anchorbase]
		\node at (0,0) {\reflectbox{\includegraphics[height=2cm]{figs/knotpicture}}};
	\end{tikzpicture}
	\longmapsto
	\begin{tikzpicture}[anchorbase]
		\node at (0,0) {\reflectbox{\includegraphics[height=2cm]{figs/knotpicture2}}};
	\end{tikzpicture}
	\longmapsto
	PD
	\longmapsto
	-q^{-4}+q^{-3}+q^{-1}
	\longmapsto
	\text{trefoil}$}
\end{gather*}
More formally: Let $\mathcal{I}$ be images at fixed resolution, $\mathcal{D}$ knot diagrams, $\mathcal{P}$ PD presentations or Morse presentations, and $\mathcal{K}$ knot types. The pipeline is the composition
\[
\Phi \;=\; \ell \circ J \circ r \circ f_{\theta} \;:\; \mathcal{I}\longrightarrow \mathcal{K},
\]
where
\begin{itemize}
	\item $f_{\theta}:\mathcal{I}\to\mathcal{D}\to\mathcal{P}$ is the NN detector (skeletonization and crossing candidates) giving a PD presentation;
	\item $r:\mathcal{P}\to\mathcal{P}$ repairs the diagram and produces a Morse normal form;
	\item $J:\mathcal{P}\to \big(\mathbb{Z}[q^{\pm1}]\big)^{k}$ computes a small tuple of invariants (e.g.\ just Jones when $k=1$);
	\item $\ell:\big(\mathbb{Z}[q^{\pm1}]\big)^{k}\to\mathcal{K}$ is a lookup/nearest-neighbor over a table of size $M$ (restricted to $c\le c_0$; ties broken by a fallback invariant).
\end{itemize}
To evaluate this, write $K$ for the ground truth knot type, $\widehat{K}$ for the output of the pipeline, and set a working threshold $c_{0}$ (e.g.\ $c_{0}\in\{12,15,20\}$) for ``small'' diagrams. Decompose
\begin{gather*}
	\Pr[\widehat{K}\neq K]
	\ \le\
	\underbrace{\varepsilon_{\mathrm{dia}}}_{\substack{\text{image $\to$ diagram}\\\text{(NN detection)}}}
	+\underbrace{\varepsilon_{\mathrm{post}}}_{\substack{\text{cleanup}\\\text{and Morse}}}
	+\underbrace{\varepsilon_{\mathrm{inv}}(c_{0})}_{\substack{\text{invariant collisions}\\\text{on }c\le c_{0}}}
	+\underbrace{\varepsilon_{\mathrm{search}}}_{\substack{\text{lookup /}\\\text{argmax tie-break}}}\!,
\end{gather*}
where:
\begin{itemize}
	\item $\varepsilon_{\mathrm{dia}}$ is the probability the detector misplaces crossings/segments enough to change isotopy type;
	\item $\varepsilon_{\mathrm{post}}$ accounts for failures in diagram repair/Morse conversion;
	\item $\varepsilon_{\mathrm{inv}}(c_{0})$ is the empirical collision rate of the selected quantum invariant(s) on $\{K: c(K)\le c_{0}\}$;
	\item $\varepsilon_{\mathrm{search}}$ covers table/database alignment and nearest-neighbor tie-breaking.
\end{itemize}
By construction, the four terms are measurable separately via ablations. 

To evaluate this strategy, the error terms $\varepsilon_{\mathrm{post}}$ and $\varepsilon_{\mathrm{search}}$ can be ignored, as they are very tiny on the knot diagrams with few crossings.
We already commented on the other two errors above:
for a tuned detector, $\varepsilon_{\mathrm{dia}}\ll 1$, and $\varepsilon_{\mathrm{inv}}(c_{0})$ can be driven to near-zero with a small invariant tuple.

Moreover, let $n$ be the number of crossings in the (repaired) diagram and $P$ the number of image pixels. Say, we compute the Jones polynomial only. The total runtime is
\begin{gather*}
	T_{\text{total}}(P,n)\ \approx O(P)\ +\ O(n)\ +\ O(2^{\sqrt{n}})\ +\ O(\log M)\approx O(2^{\sqrt{n}}),
\end{gather*}
where $O(P)$ is one forward pass, $O(n)$ covers repair/Morse conversion, $O(2^{\sqrt{n}})$ is the invariant computation (feasible for $n\le c_{0}$ in our setup), and $M$ is the size of the lookup table. For $n\le c_{0}\le 20$ (or $30$ etc.), $O(2^{\sqrt{n}})$ is the bottleneck but remains practical, and the $O(\log M)$ search is negligible.


\subsection{Optional extension to 3D images}

We extend the pipeline from single images to multi-view / video inputs, leveraging depth cues and temporal consistency to reduce over/under ambiguity and improve robustness. 

To this end, let $\mathcal{V}$ denote videos or multi-view sets (ordered frames), and let $\mathcal{C}$ denote embedded $C^1$ space curves (centerlines) in $\mathbb{R}^3$. We introduce a reconstruction map to a 3D curve and a projection-selection step before diagramming:
\[
\Phi_{\mathrm{3D}}
\;=\;
\ell \circ J \circ \Pi \circ r \circ D \circ F \circ f_{\theta}
\;:\;
\mathcal{V}\longrightarrow \mathcal{K},
\]
where:
\begin{itemize}
  \item $F:$ multi-frame association and fusion (tracking) $\to$ a 3D centerline $\Gamma\in\mathcal{C}$ (via stereo/N-view triangulation or monocular depth + temporal smoothing);
  \item $D:$ candidate crossing detection on projections. Outputs labeled junctions and preliminary PD per view;
  \item $\Pi:$ choose a generic projection $\pi:\mathbb{R}^3\!\to\!\mathbb{R}^2$ (or a small set $\{\pi_j\}$) that minimizes diagram complexity (e.g.\ estimated crossing count) and stabilizes crossing signs (depth ordering). 
\end{itemize}

Why would 3D/video help?
Depth and time provide two powerful regularizers that the 2D case lacks:
\begin{enumerate}[resume]
  \item Depth disambiguation. Over/under at a detected crossing is $\operatorname{sign}(\Delta z)$ after projecting by $\pi$; multi-view disparity or reliable monocular depth reduces label flips.
  \item Temporal coherence. Crossing identities can be tracked across nearby frames; labels should vary only by admissible Reidemeister updates. This yields a global consistency prior over the PD presentation. 
\end{enumerate}


The error decomposition is now as follows.
Let $K$ be the ground truth, $\widehat{K}$ the output. For a working threshold $c_0$ as before,
\begin{gather*}
\Pr[\widehat{K}\neq K]\ \le\
\underbrace{\varepsilon_{\mathrm{seg}}}_{\text{per-frame rope masks/skeletons}}
+\underbrace{\varepsilon_{\mathrm{trk}}}_{\text{temporal association}}
+\underbrace{\varepsilon_{\mathrm{3D}}}_{\text{curve triangulation / depth}}
+\underbrace{\varepsilon_{\mathrm{proj}}}_{\text{projection choice}} \\
+\underbrace{\varepsilon_{\mathrm{post}}}_{\text{repair/Morse}}
+\underbrace{\varepsilon_{\mathrm{inv}}(c_0)}_{\text{invariant collisions on }c\le c_0}
+\underbrace{\varepsilon_{\mathrm{search}}}_{\text{lookup/tie-break}}.
\end{gather*}
All terms are separately measurable via ablations:
\begin{itemize}
  \item $\varepsilon_{\mathrm{seg}}$ from mask/centerline IoU and junction recall,
  \item $\varepsilon_{\mathrm{trk}}$ from identity switches and track purity,
  \item $\varepsilon_{\mathrm{3D}}$ from 3D reprojection error and curve smoothness,
  \item $\varepsilon_{\mathrm{proj}}$ by evaluating a shortlist $\{\pi_j\}$ and checking diagram consistency/complexity,
  \item $\varepsilon_{\mathrm{post}},\varepsilon_{\mathrm{inv}}(c_0),\varepsilon_{\mathrm{search}}$ as in the 2D setting.
\end{itemize}
Depth and temporal voting typically shrink the dominant 2D term (over/under flips) by aggregating consistent evidence across frames/views. With near-independent per-view crossing labels of error rate $p<\tfrac12$, majority voting over $T$ views yields a binomial tail bound $\lesssim e^{-2(1-2p)^2 T}$ for each crossing (heuristic but indicative).

To analyze the runtime, let $P$ be pixels per frame, $T$ frames, and let $n^\ast$ be the crossing count of the selected projection (expected: $n^{\ast}\le c_0$ due to $\Pi$).
\[
T_{\text{total}}
\;\approx\;
\underbrace{O(TP)}_{\text{NN passes}}
+\underbrace{O(T\,\bar n)}_{\text{track/fuse/repair}}
+\underbrace{O\!\big(J\,(K P + K\,\bar n)\big)}_{\text{projection scoring}}
+\underbrace{O\!\big(2^{\sqrt{n^{\ast}}}\big)}_{\text{invariants on best view}}
+\underbrace{O(\log M)}_{\text{lookup}} \,.
\]
Here $\bar{n}$ is the typical per-frame candidate crossing count. Using $K\ll T$ keyframes or a small candidate set $\{\pi_j\}$ keeps $n^\ast$ small; for $n^\ast\le c_0\in\{20,30\}$, $O(2^{\sqrt{n^\ast}})$ remains practical and dominates, while $O(\log M)$ is negligible.


A practical recipe could be:
\begin{enumerate}[resume]
  \item Detect \& skeletonize each frame (DLO segmentation).
  \item Associate centerline pieces temporally; fuse to a smooth $\Gamma$ (stereo/N-view if available).
  \item Generate a small set of generic projections $\{\pi_j\}$; score by estimated crossing count and label stability (depth ordering).
  \item Build PD for top-scoring $\pi_j$, repair to Morse, compute $J$, then lookup $\ell$ restricted to $c\le c_0$; break ties with a fallback invariant or the braid branch if applicable.
\end{enumerate}

\begin{Remark}
Compared to the 2D image recipe, the additional steps (temporal fusion, projection selection) exploit the extra signals of depth and time. This keeps the pipeline mathematically principled (explicit isotopy-preserving steps; invariants computed on valid diagrams) and computationally feasible in the small-crossing regime, while reducing the dominant failure mode of the 2D case: over/under misclassification.
\end{Remark}

\section{Crossing detection: vanilla versus CNN versus transformer}\label{S:CrossingDetection}

In what follows, we employ three types of NN architectures: two classical ones and one inspired by \cite{Wu2021CvT}. Specifically, we use a multilayer perceptron (MLP, hereafter Vanilla), a convolutional neural network (CNN), and a convolutional transformer (CvT).

\subsection{Data preparation}

We now describe how raw images were normalized into inputs suitable for our models. The guiding principle is to expose the combinatorics of crossings while suppressing photometric and geometric nuisance variables.

We began with knot images from \cite{Ri20} (henceforth, Prideout images) as they are very clear. Since the source format is \texttt{.svg}, we rasterized each file to \texttt{.png} and standardized the resolution to $512\times512$ pixels so that small crossings remain well resolved. Subsequent I/O was handled with Python’s Pillow (\texttt{PIL}) library (via \texttt{Image}) within our \texttt{PyTorch} pipeline.

\begin{figure}[ht]
\centering
  \begin{subfigure}[b]{0.44\textwidth}\centering
    \includegraphics[height=4cm]{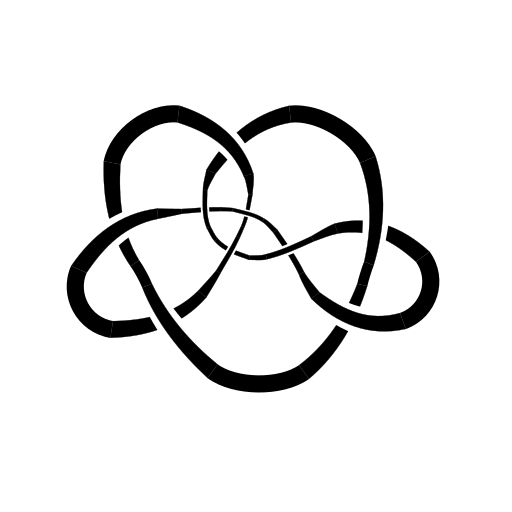}
    \caption{Original image.}
  \end{subfigure}\hfill
  \begin{subfigure}[b]{0.44\textwidth}\centering
    \includegraphics[height=4cm]{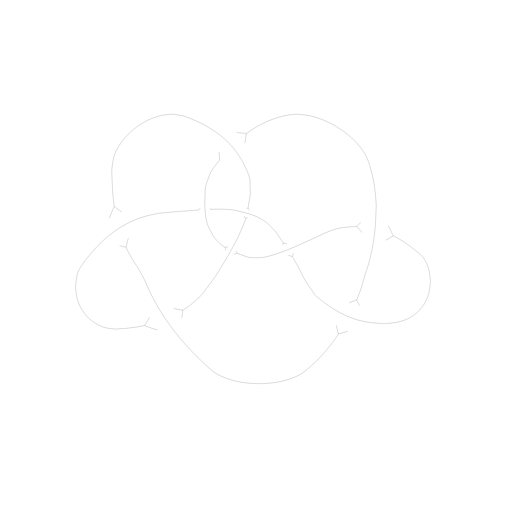}
    \caption{Skeletonized image.}
  \end{subfigure}
  \caption{Data preparation: from the original raster to a skeletonized diagram.}
  \label{Fig:dataprep}
\end{figure}

Our aim is to represent a planar knot diagram by its 1-pixel centerline and crossing structure, not strand thickness. We therefore converted to grayscale and applied topology-preserving skeletonization (via \texttt{scikit-image}) to remove width while retaining the degree-4 junctions that encode crossings. In short, the initial preparation was:
\begin{enumerate}[resume]
  \item Rasterize the Prideout \texttt{.svg} to a $512\times512$ \texttt{.png}.
  \item Convert to grayscale and skeletonize.
  \item Save the processed image.
\end{enumerate}
See \autoref{Fig:dataprep} for an illustration.

\begin{figure}[ht]
\centering
  \begin{subfigure}[b]{0.44\textwidth}\centering
    \includegraphics[height=4cm]{figs/knot_skeletonize.png}
    \caption{Skeletonized.}
  \end{subfigure}\hfill
  \begin{subfigure}[b]{0.44\textwidth}\centering
    \includegraphics[height=4cm]{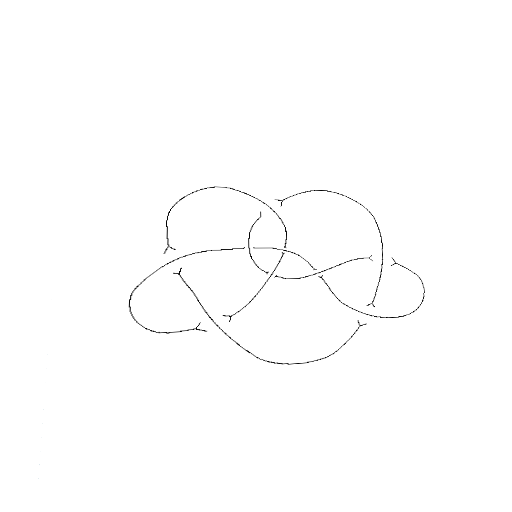}
    \caption{After perspective transform.}
  \end{subfigure}\\[0.6em]
  \begin{subfigure}[b]{0.44\textwidth}\centering
    \includegraphics[height=4cm]{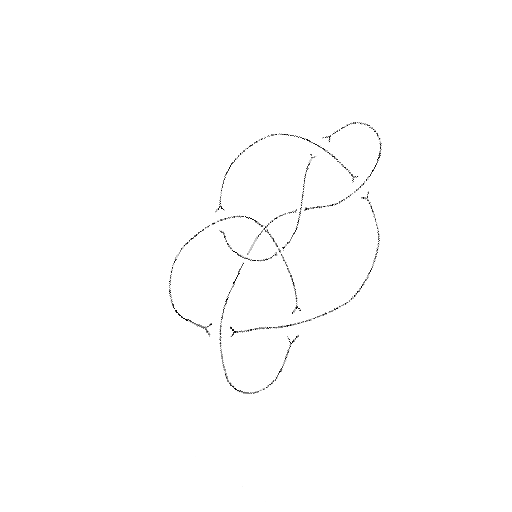}
    \caption{Random flips/rotations.}
  \end{subfigure}\hfill
  \begin{subfigure}[b]{0.44\textwidth}\centering
    \includegraphics[height=4cm]{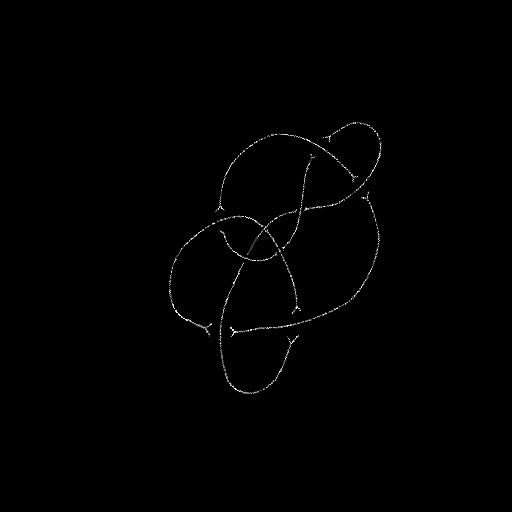}
    \caption{Inverted and binarized.}
  \end{subfigure}
  \caption{Data augmentation: geometric variations followed by binarization.}
  \label{Fig:finaldata}
\end{figure}

To increase diversity and reduce spurious symmetries, we augmented each skeletonized image as follows:
\begin{enumerate}[resume]
  \item Apply a random perspective transform (to emulate viewpoint changes and break global symmetries).
  \item Apply random horizontal/vertical flips and random rotations (distributed evenly across the dataset).
  \item Adjust per crossing number so that classes remain balanced for our task (predicting the crossing count; e.g.\ $6_1$ and $6_2$ are indistinguishable for this purpose).
  \item Invert and binarize (foreground/knot pixels set to $1$, background to $0$) to present a clean signal to the models.
\end{enumerate}
This is illustrated in \autoref{Fig:finaldata}.

The final corpus comprises $9051$ images, with a held-out test set of $1811$ images.




\subsection{Training process}

We now explain how the networks were trained. A key observation is that predicted crossing numbers admit a natural notion of distance: for example, mispredicting a $4_1$ knot as having $5$ crossings is far less severe than predicting $10$. This motivates the use of a regression loss rather than a strict classification loss. Concretely, we employed mean squared error (MSE) loss, which directly penalizes deviations in the predicted crossing count. As a consequence, the problem shifts from a multiclass classification task to a regression task, and each model outputs a real number representing the estimated crossing count. This also makes the models more robust for generalization to knots with larger crossing numbers than those present in the training set.

All models were trained with the AdamW optimizer under different schedules of epochs, learning rates, and weight decay. The details are summarized in \autoref{Tab:trainopt}.

\begin{table}[ht]
\centering
\begin{tabular}{c|c|c|c|c|c|c}
Model & Epochs & Optimizer & Learning rate & Weight decay & Loss & Scheduler \\ \hline
Vanilla & 20  & AdamW & 0.0003 & 0.0001 & MSELoss & ReduceLROnPlateau \\
CNN & 30  & AdamW & 0.0003 & 0.0001 & MSELoss & ReduceLROnPlateau \\
CvT & 70  & AdamW & 0.003  & 0.01   & MSELoss & ReduceLROnPlateau \\
\end{tabular}
\caption{Training hyperparameters used for different runs.}
\label{Tab:trainopt}
\end{table}

The learning rate scheduler was \texttt{ReduceLROnPlateau}, with model-specific parameters for factor, patience, and cooldown. These values are listed in \autoref{Tab:sched}.

\begin{table}[ht]
\centering
\begin{tabular}{c|c|c|c}
Model   & Factor & Patience & Cooldown \\ \hline
Vanilla & 0.2    & 3        & 0 \\
CNN     & 0.2    & 7        & 0 \\
CvT     & 0.7    & 1        & 7 \\
\end{tabular}
\caption{Scheduler parameters for different architectures.}
\label{Tab:sched}
\end{table}

\subsection{Architectures}

We now describe the three NN architectures used in our experiments: Vanilla, CNN, and CvT. Each is summarized below together with its parameter counts.

\subsubsection{Vanilla}

The Vanilla model is a multilayer perceptron with six layers (zero-indexed). The input layer (layer 0) applies \texttt{Flatten} followed by \texttt{BatchNorm1d}. Each subsequent layer consists of a linear transformation, an ELU activation, and \texttt{BatchNorm1d}. Before the final layer we inserted a dropout layer with rate $0.9$, though this did not yield noticeable benefits.

The parameter counts per layer are given in \autoref{Tab:vanilla}. In total the model contains $127{,}807{,}135$ trainable parameters.

\begin{table}[ht]
\centering
\begin{tabular}{c|c}
Layer & Number of parameters \\ \hline
1 & $262{,}144 \cdot 484 = 126{,}877{,}696$ \\
2 & $484 \cdot 576 = 278{,}784$ \\
3 & $576 \cdot 196 = 112{,}896$ \\
4 & $196 \cdot 49 = 9{,}604$ \\
5 & $49 \cdot 1 = 49$ \\
\end{tabular}
\caption{Parameter counts per layer for the Vanilla model.}
\label{Tab:vanilla}
\end{table}

\subsubsection{CNN}

The CNN architecture consists of a stack of convolutional blocks followed by a linear layer. Each convolutional block has the structure
\[
\texttt{Conv2d} \;+\; \texttt{BatchNorm2d} \;+\; \texttt{ReLU} \;+\; \texttt{MaxPool2d},
\]
where all pooling layers use kernel size $2$. The specific convolutional parameters are listed in \autoref{Tab:cnn}.  

\begin{table}[ht]
\centering
\begin{tabular}{c|c|c|c|c|c|c}
Block & In-channels & Out-channels & Kernel size & Stride & Dilation & Padding \\ \hline
1 & 1   & 4   & 16 & 1 & 2 & 0 \\
2 & 4   & 16  & 6  & 1 & 2 & 0 \\
3 & 16  & 64  & 4  & 1 & 2 & 0 \\
4 & 64  & 256 & 3  & 1 & 2 & 0 \\
5 & 256 & 256 & 3  & 1 & 0 & 0 \\
\end{tabular}
\caption{Convolutional layer parameters for the CNN model.}
\label{Tab:cnn}
\end{table}

The final linear layer flattens the feature map and applies a fully connected layer, with $(256 \cdot 121)\cdot 1 = 30{,}976$ parameters. In total, the CNN has $230{,}820{,}732$ parameters, approximately $1.8$ times more than the Vanilla model.

\subsubsection{CvT}

\begin{figure}[ht]
\centering
\includegraphics[height=6cm]{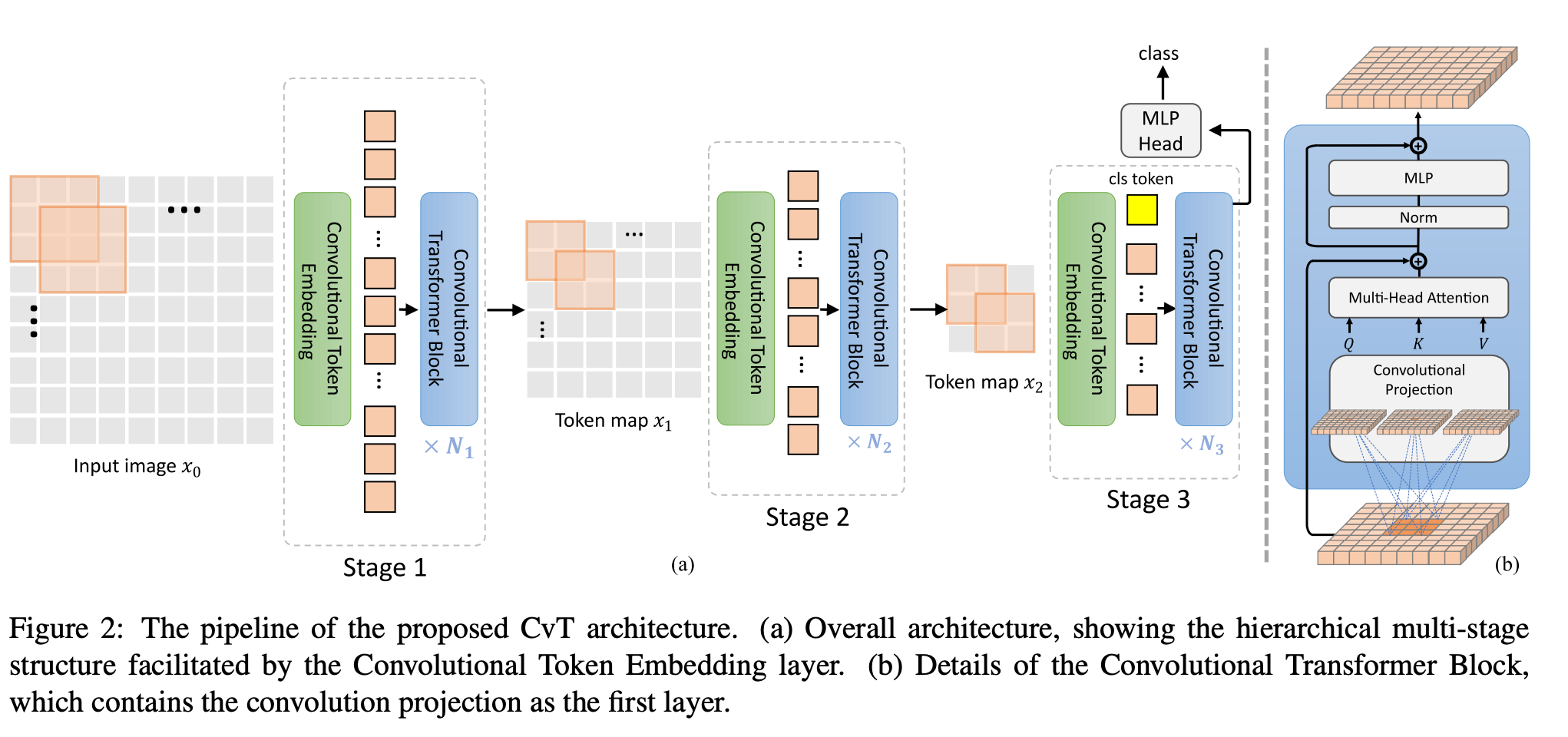}
\caption{Schematic of the CvT architecture (as in \cite{Wu2021CvT}).}
\label{Fig:cvtarch}
\end{figure}

The third model is a convolutional transformer (CvT), following the design proposed in the original paper \cite{Wu2021CvT}. Its high-level architecture is shown in \autoref{Fig:cvtarch}.  

We implemented three stages consisting of $2$, $3$, and $9$ ConvTransformer blocks, respectively. The parameters of each stage are summarized in \autoref{Tab:cvt}.  

\begin{table}[ht]
\centering
\begin{tabular}{c|c|c|c|c|c|c|c|c}
Layer & $N_s$ & Patch size & Stride & In-dim & Embed-dim & Heads & MLP-dim & Out-dim \\ \hline
1 & 2 & 32 & 32 & 1  & 32  & 8 & 256 & 32 \\
2 & 3 & 8  & 4  & 32 & 64  & 8 & 256 & 64 \\
3 & 9 & 3  & 1  & 64 & 128 & 8 & 256 & 128 \\
\end{tabular}
\caption{Stage parameters for the CvT model. $N_s$ is the number of ConvTransformer blocks.}
\label{Tab:cvt}
\end{table}

In total, the CvT has $3{,}210{,}273$ parameters, significantly fewer than either the CNN or the Vanilla model. The main limitation is memory usage: the attention mechanisms introduce dense dependencies that make training memory-intensive. This imposes a trade-off: too small a model risks vanishing gradients and insufficient capacity, while too large a model exceeds feasible memory limits. We found the chosen configuration to balance these constraints reasonably well.

\subsubsection{Summary}

Taken together, the three architectures span a spectrum of design choices. The Vanilla model is parameter-heavy yet structurally simple, relying purely on dense connections. The CNN introduces spatial locality and achieves better efficiency at the cost of a larger parameter count. The CvT, finally, incorporates attention mechanisms that drastically reduce the number of parameters, but at the price of higher memory consumption and training instability. This trade-off between size, expressivity, and computational feasibility underpins our choice to evaluate all three side by side.

\subsection{Overall results}

\begin{figure}[ht]
\centering
\includegraphics[height=6cm]{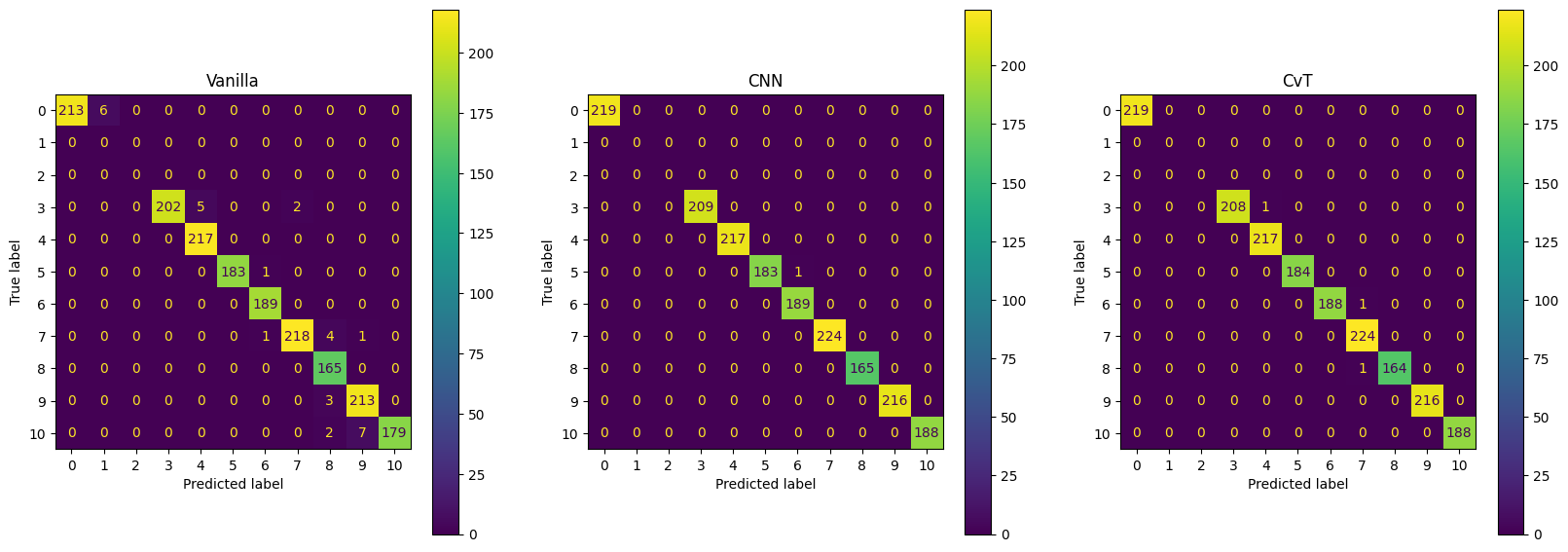}
\caption{Confusion matrices for the three models: Vanilla, CNN, and CvT.}
\label{Fig:confmatrices}
\end{figure}

We now compare the performance of the three models. Both the CNN and the CvT outperform the Vanilla baseline by roughly $1.5\%$ in accuracy. Their confusion matrices are shown in \autoref{Fig:confmatrices}.

\begin{figure}[ht]
\centering
\includegraphics[height=6cm]{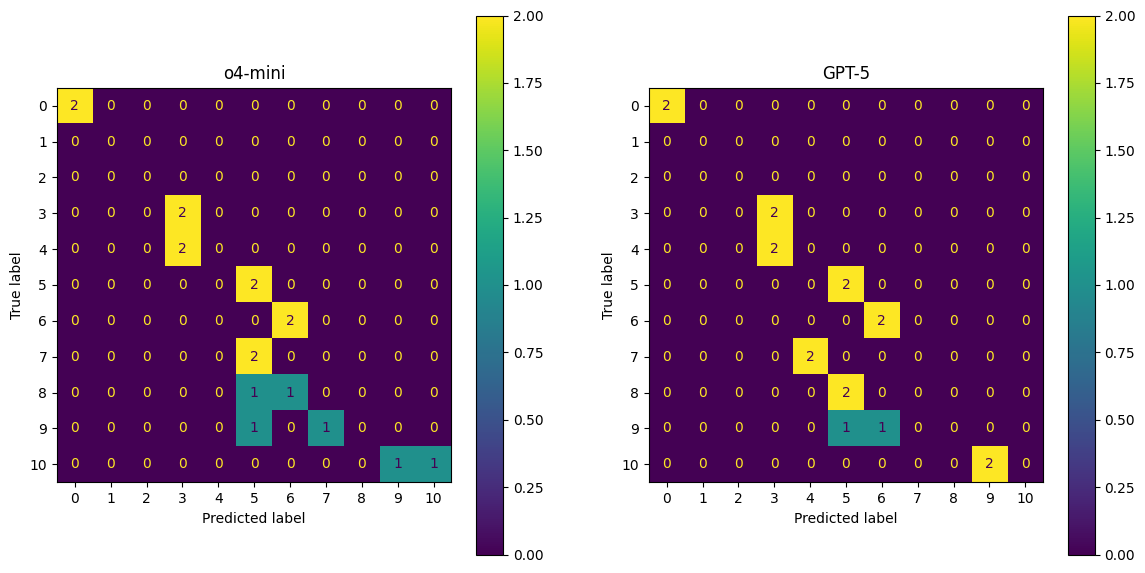}
\caption{Confusion matrix for the OpenAI \texttt{o4-mini} and \texttt{GPT-5} baseline.}
\label{Fig:confmatrgpt}
\end{figure}

For reference, we also evaluated the OpenAI \texttt{o4-mini} and \texttt{GPT-5}. The confusion matrix is given in \autoref{Fig:confmatrgpt}. Although the test set is small, even coarse comparison shows that both models perform substantially worse. \texttt{o4-mini} has one more correct guessing than \texttt{GPT-5}, so one can say that both results are actually the same.

\begin{table}[ht]
\centering
\begin{tabular}{c|c}
Model   & Accuracy \\ \hline
o4-mini & $50\%$ \\
GPT-5 & $\approx 44.44\%$ \\
Vanilla & $\approx 98.23\%$ \\
CNN     & $\approx 99.94\%$ \\
CvT     & $\approx 99.83\%$ \\
\end{tabular}
\caption{Accuracy comparison across models.}
\label{Tab:results}
\end{table}

\begin{figure}[ht]
\centering
\includegraphics[height=6cm]{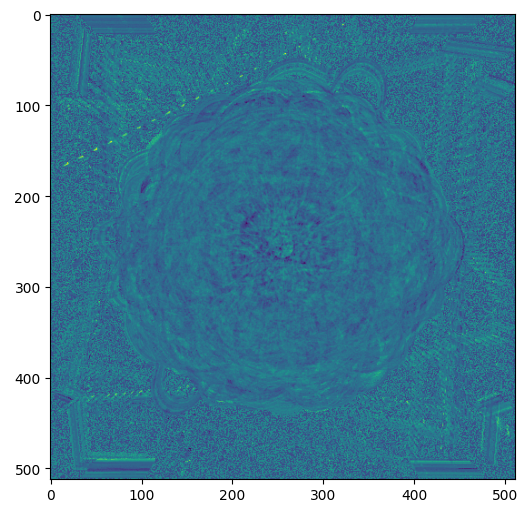}
\caption{Visualization of the first layer weights in the Vanilla network.}
\label{Fig:weightsvanilla}
\end{figure}

The gap between Vanilla and the other two models highlights the importance of architectures that exploit spatial structure. The Vanilla network processes pixels as a flattened vector. It therefore fails to capture geometric locality and instead learns a distributional representation of training pixels. This suffices for data drawn from the same style as the training set but generalizes poorly. Evidence for this is provided by the weight distribution in the first layer (see \autoref{Fig:weightsvanilla}), which shows little interpretable structure.

\begin{figure}[ht]
\centering
  \begin{subfigure}[b]{0.44\textwidth}\centering
    \includegraphics[height=4cm]{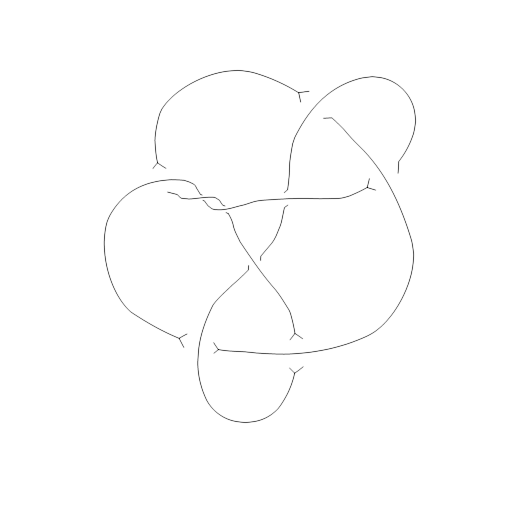}
    \caption{$9_{46}$}
  \end{subfigure}\hfill
  \begin{subfigure}[b]{0.44\textwidth}\centering
    \includegraphics[height=4cm]{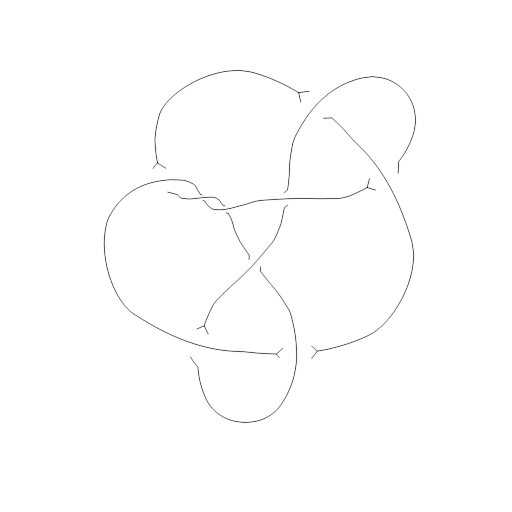}
    \caption{$9_{48}$}
  \end{subfigure}
\caption{Two visually similar knots, $9_{46}$ and $9_{48}$, that CvT partially distinguishes.}
\label{Fig:sameknots}
\end{figure}

By contrast, the CNN and CvT both incorporate spatial priors and attention mechanisms, allowing them to capture crossing structure robustly. On the test set, the CNN misclassified only one knot, while the CvT misclassified two. Interestingly, the CvT occasionally distinguishes between visually similar knots (such as $9_{46}$ and $9_{48}$) by attending to localized features of the diagram, cf.\ Figure~\ref{Fig:sameknots}.

The CNN converged in roughly $30$ epochs, while the CvT required $70$. This discrepancy reflects the fact that in each epoch the CvT learns less useful information, owing to its heavier reliance on attention across the entire image rather than purely local filters.

\begin{figure}[ht]
\centering
  \begin{subfigure}[b]{0.32\textwidth}\centering
    \includegraphics[height=3.8cm]{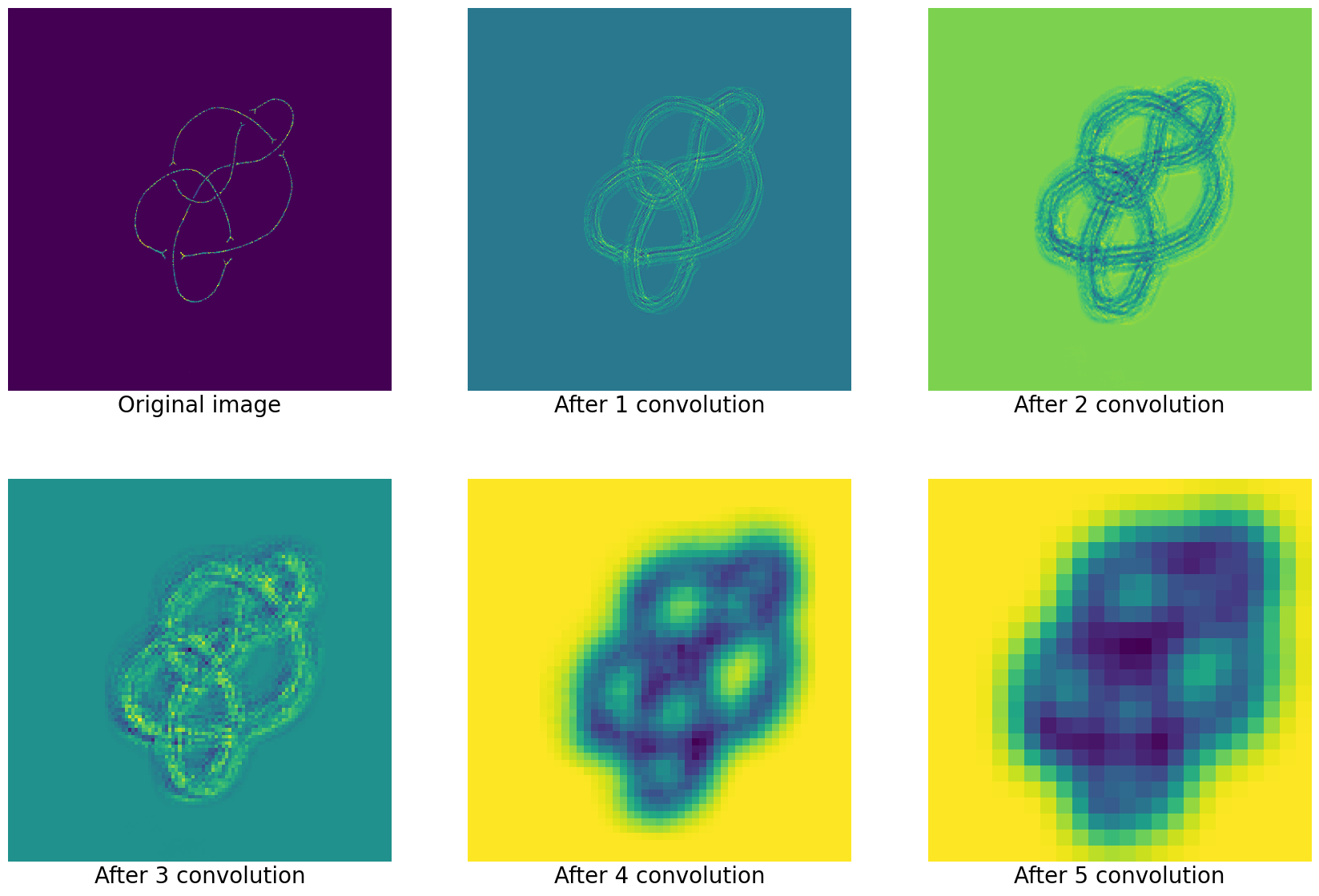}
    \caption{9 crossings}
  \end{subfigure}
  \begin{subfigure}[b]{0.32\textwidth}\centering
    \includegraphics[height=3.8cm]{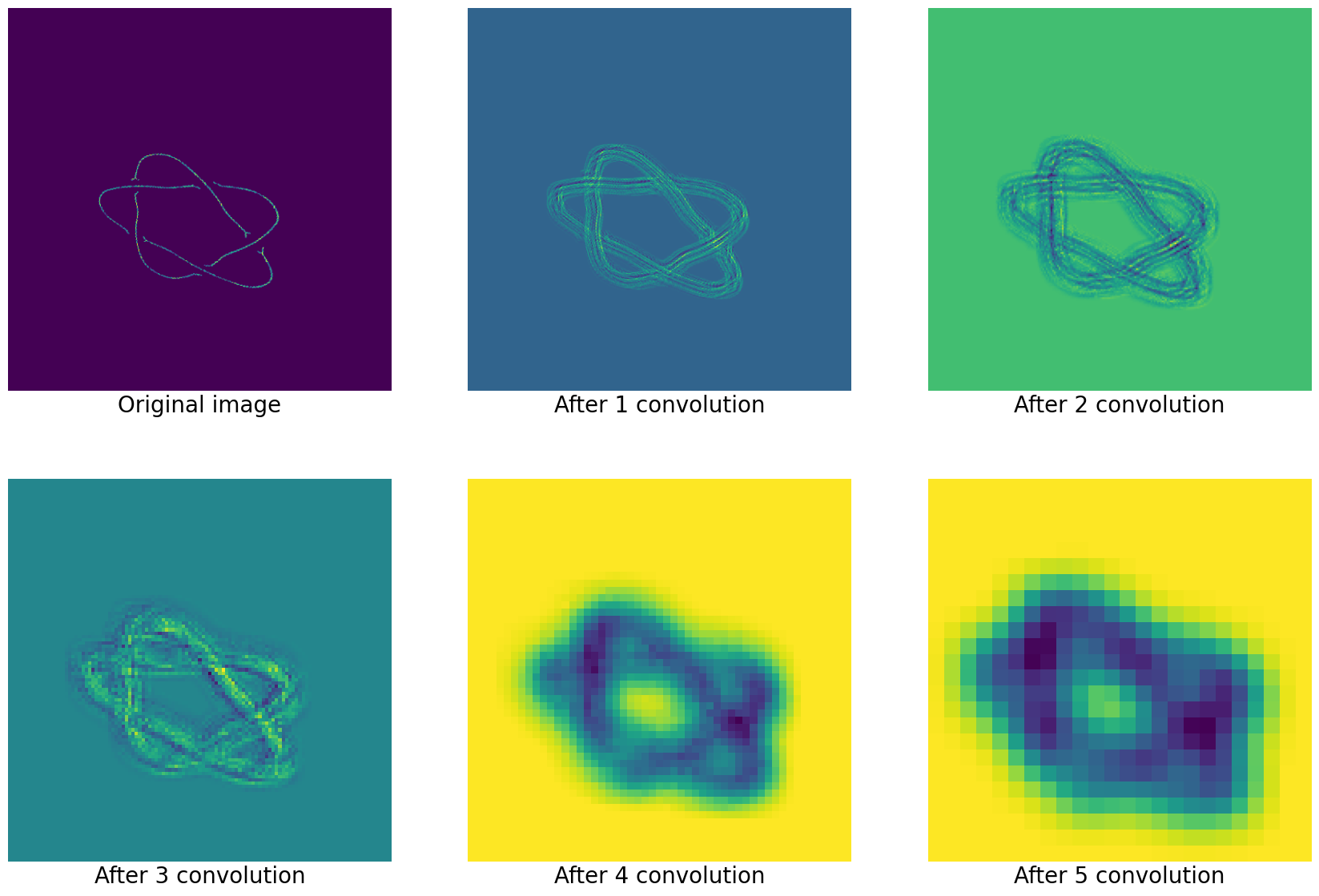}
    \caption{5 crossings}
  \end{subfigure}
  \begin{subfigure}[b]{0.32\textwidth}\centering
    \includegraphics[height=3.8cm]{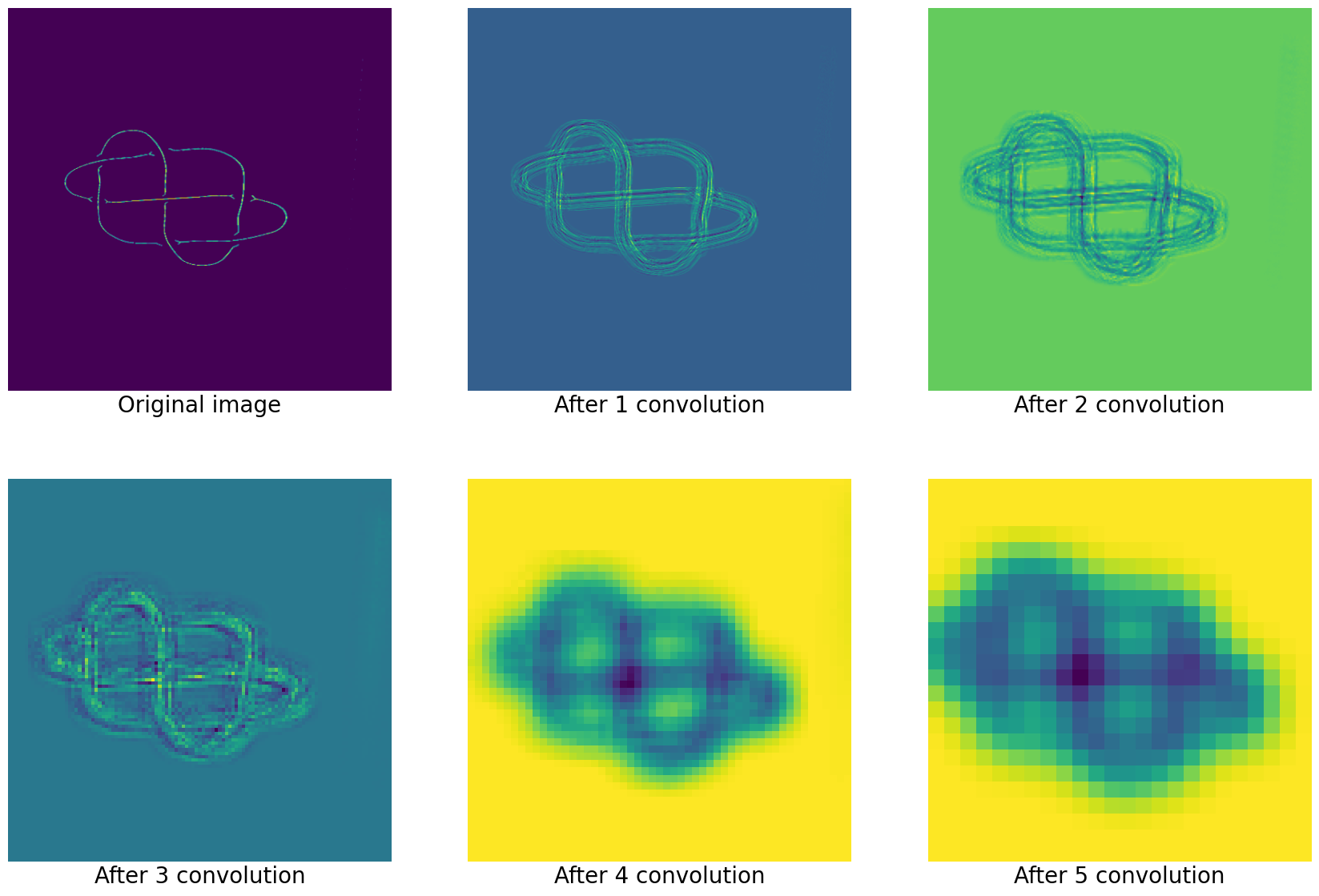}
    \caption{7 crossings}
  \end{subfigure}
\caption{Intermediate feature maps in the CNN.}
\label{Fig:cnnexs}
\end{figure}

\begin{figure}[ht]
\centering
  \begin{subfigure}[b]{0.32\textwidth}\centering
    \includegraphics[height=4cm]{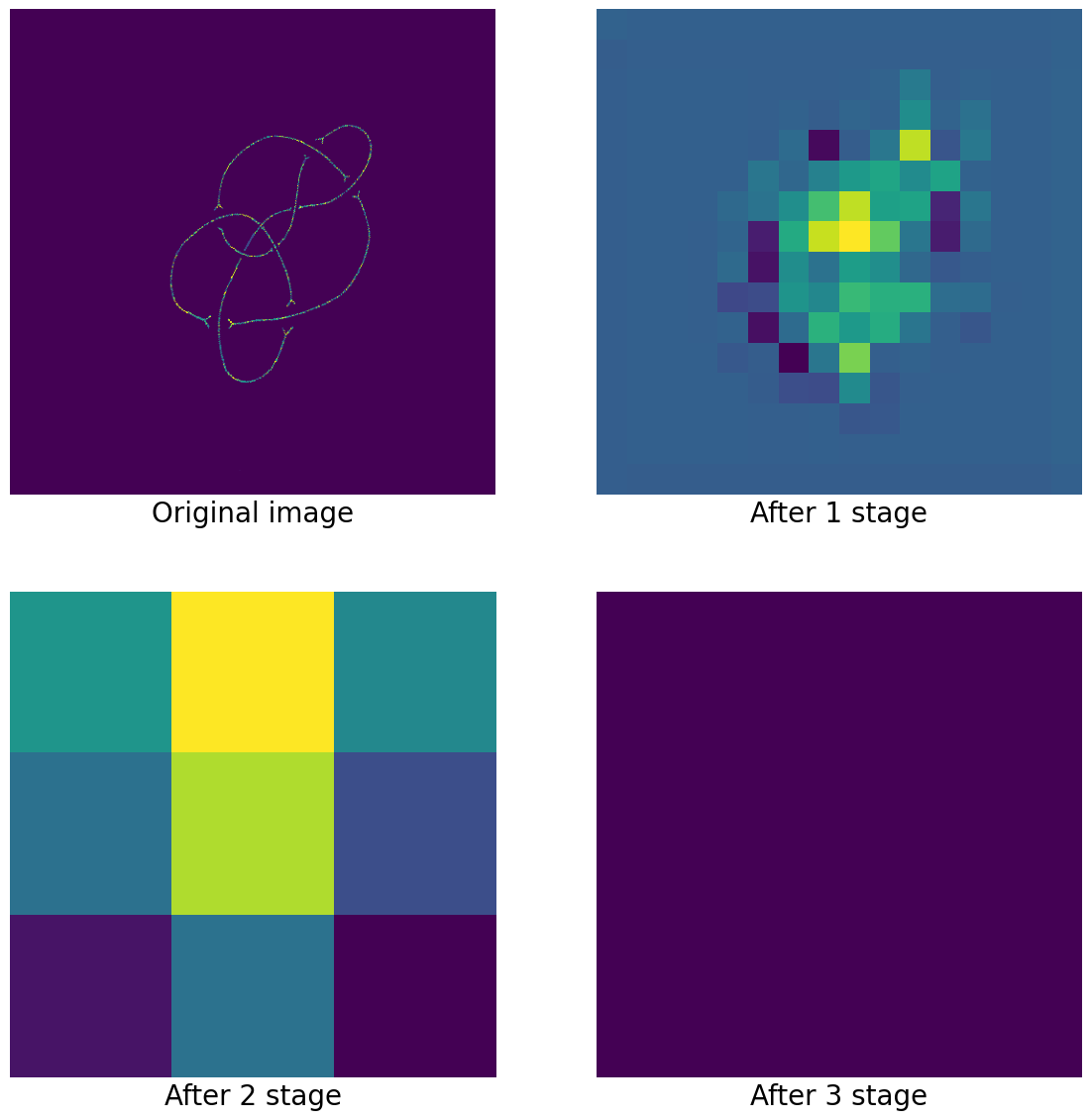}
    \caption{9 crossings}
  \end{subfigure}
  \begin{subfigure}[b]{0.32\textwidth}\centering
    \includegraphics[height=4cm]{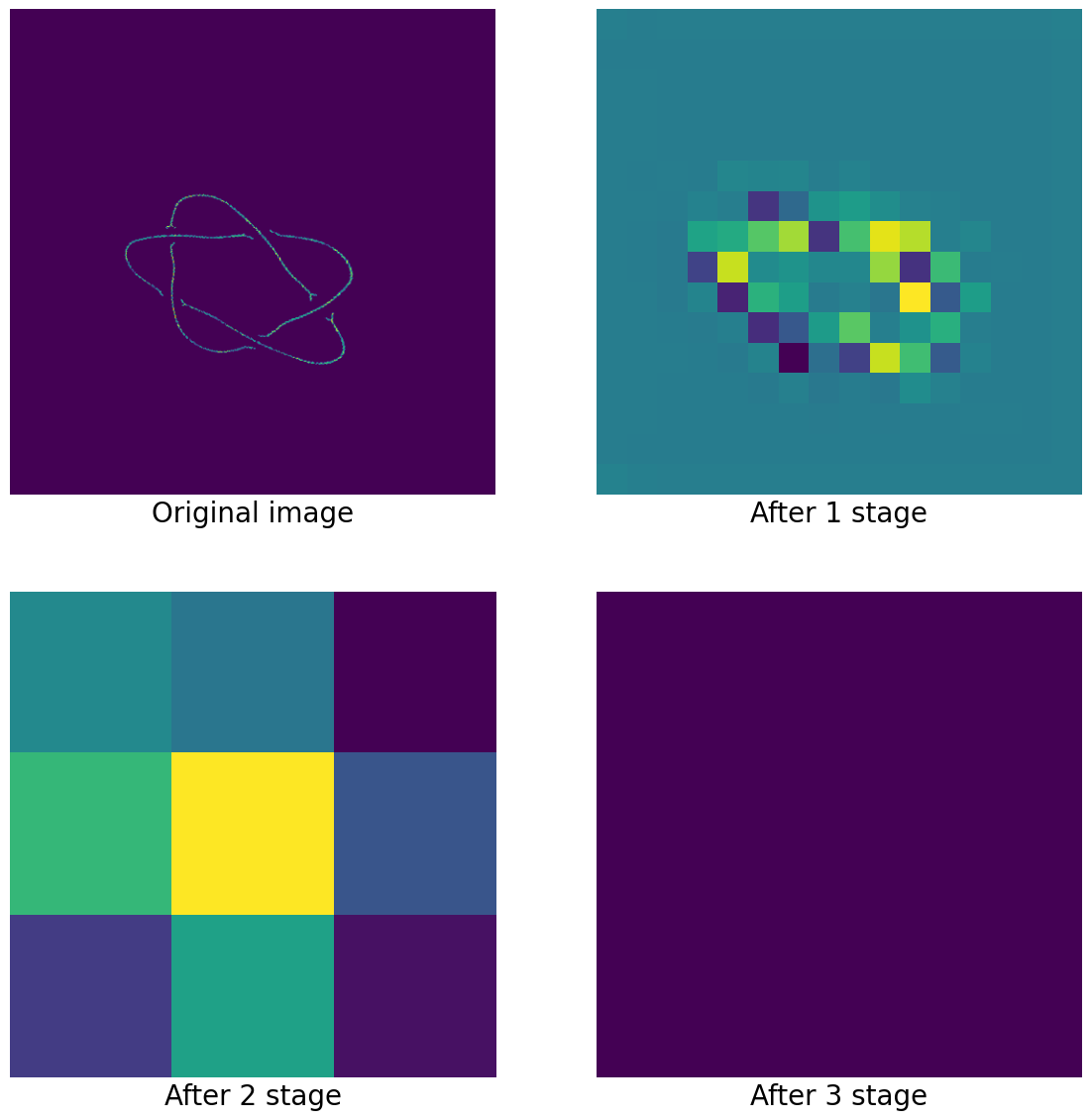}
    \caption{5 crossings}
  \end{subfigure}
  \begin{subfigure}[b]{0.32\textwidth}\centering
    \includegraphics[height=4cm]{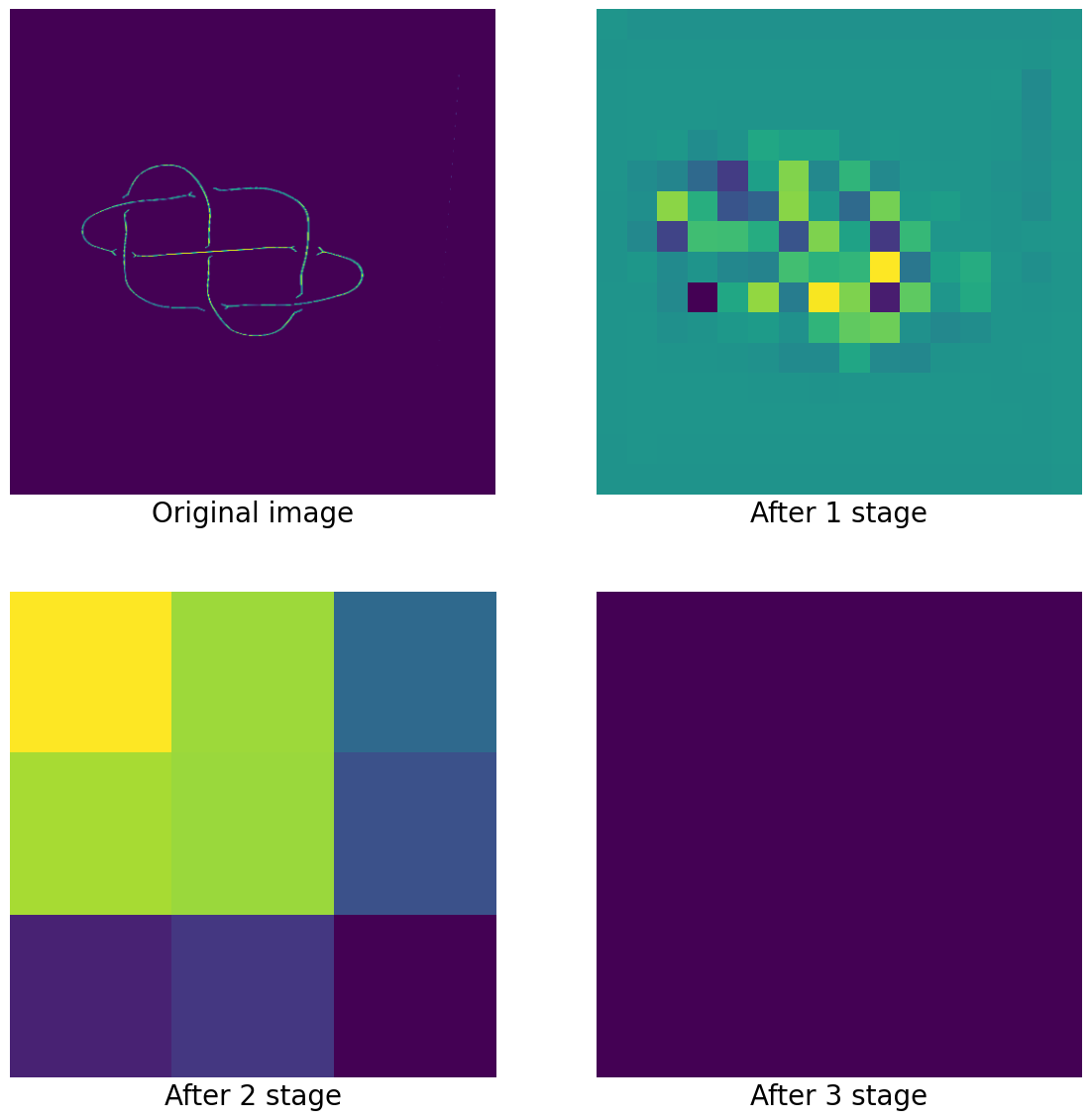}
    \caption{7 crossings}
  \end{subfigure}
\caption{Intermediate feature maps in the CvT.}
\label{Fig:cvtexs}
\end{figure}

\autoref{Fig:cnnexs} and \autoref{Fig:cvtexs} illustrate the intermediate activations of the CNN and CvT, respectively. The CNN progressively extracts crossings and maps them to a scalar crossing count. The CvT, in contrast, tends to distribute attention across patches, effectively “pixelizing” the knot, but preserves enough information to recover the crossing number. In both models, the final step is a linear layer acting as a weighted sum. This highlights a general trade-off: deeper convolutional hierarchies tend to be more efficient than attention-heavy models under the same memory budget, but the latter provide more flexibility in capturing global patterns.

In summary, all three models achieve strong performance on the small-knot regime, but architectures that explicitly exploit spatial structure clearly outperform dense baselines. The CNN achieves the best balance of accuracy and efficiency, while the CvT demonstrates the potential of attention mechanisms to capture subtle global patterns, albeit at higher memory cost. These findings support our overall strategy: in the context of visual knot recognition, architectures with built-in geometric inductive bias are both more accurate and more robust to generalization.




\subsection{What the models actually learn}\label{subsec:interpretation}

The CNN builds edge/vertex detectors that coalesce into ``crossing counters'' by depth; its final linear head then acts as a smooth aggregator of localized crossing evidence. The CvT attends to patchwise structures that effectively ``pixelize'' the diagram yet preserve enough global consistency to infer crossing counts. In contrast, the Vanilla model treats the image as a long vector; its early weights show little interpretable structure, and it appears to memorize dataset style rather than geometry.

The residual errors of CNN/CvT are consistent with three sources: (i) borderline diagrams where augmentation slightly shifts perceived local geometry, (ii) rare topology–preserving artifacts of skeletonization, and (iii) outliers whose visual style deviates from the training distribution.
The near‐saturation accuracies reflect that our data are skeletonized, style‐controlled diagram images. Under domain shift (hand‐drawn diagrams, Rolfsen‐style renderings, photographs with specularities, partial occlusion), we expect a drop unless the database is diversified. The inductive bias ranking we observe CNN~$\gtrsim$~CvT~$\gg$~Vanilla and this should persist, but the absolute margins will depend on how well the training distribution covers line quality, stroke width, and projection artifacts.

However, estimating the crossing count is largely a \emph{local} task: small receptive fields already capture the discriminative cues (junction geometry and short-range strand continuations), which explains the CNN’s advantage. In contrast, recovering a consistent planar diagram (a PD presentation) and enforcing over/under coherence are inherently \emph{global}: long-range dependencies and multi-patch consistency matter. Hence, one should expect self-attention to shine on PD-presentation-level objectives (or hybrids that add attention on top of a convolutional backbone), even if CNNs dominate the crossing-count surrogate.

\subsection{Future improvement}

While our initial experiments demonstrate that CNNs and CvTs can achieve near-perfect accuracy on small knots, there remain several avenues for further improvement. We highlight two directions here: enlarging the training database and exploring alternative architectures.

\subsubsection{Database augmentation} 

A straightforward way to improve performance is to enlarge and diversify the training database. Our CvT implementation was intentionally small, and reducing its parameter count any further led to sharp drops in accuracy and increases in loss. At the same time, both the CNN and CvT were trained on datasets of relatively uniform style (see our GitHub repository for details). As a result, they do not generalize well to more heterogeneous examples, such as those from the Rolfsen table. Expanding the dataset with diagrams of varied style and complexity is therefore a natural next step.

\subsubsection{Different architectures}

Another promising avenue is to explore alternative network architectures. For example, convolutional kernels with differentiable size could help address one of the key challenges in this setting: finding the right balance between memory consumption and information retention. More broadly, there is currently no implementation of a differentiable convolutional transformer, but such an architecture may be particularly well-suited for knot diagram recognition. Designing and testing these variants would provide valuable insights into the trade-offs between efficiency, expressivity, and stability.

Together, these improvements would strengthen the overall pipeline from image to planar diagram to invariant, extending its applicability beyond small synthetic examples and bringing us closer to robust, automated knot recognition in realistic settings.



\subsubsection{Application to DNA knots}\label{subsec:4E3-DNA}

DNA molecules often form knots during replication, recombination, and viral packaging. An example of an open dataset is \cite{Szumilo2024DNAKnots}, provides raw and processed atomic force microscopy images of knotted and catenated DNA molecules. This collection is ideal for our purposes: most observed knots are of low crossing number and dominated by torus families, making them ``easy'' cases in our pipeline. By augmenting our training data with these AFM images, we can directly test the robustness of our image $\longmapsto$ PD presentation $\longmapsto$ invariant approach on experimentally observed DNA structures.

\subsubsection{Application to protein knots}\label{subsec:4E4-Proteins}

Knots also occur in proteins, though more rarely than in DNA, with a marked bias toward certain easy knots such as twist knots. The main curated resources are the database \cite{Jamroz2015KnotProt} and its extension \cite{Jarmolinska2024AlphaKnot}, which compile all known knotted proteins (including AlphaFold predictions) and annotate them by type. These provide ready-made training and benchmarking sets: backbone curves can be projected to diagrams and classified via our invariant lookup, enabling efficient large-scale annotation of knotted proteins.


\newcommand{\etalchar}[1]{$^{#1}$}

\end{document}